\documentclass[final,5p,times,twocolumn,a4paper,margin=1in]{elsarticle}
\usepackage[colorlinks=true,linkcolor=blue,anchorcolor=blue,citecolor=blue,filecolor=blue,menucolor=blue,runcolor=blue,urlcolor=blue,breaklinks]{hyperref}
\usepackage{subcaption}
\usepackage{float}
\usepackage{amsmath}
\usepackage{amsthm}
\usepackage{algorithm}
\usepackage{float}
\usepackage{booktabs}
\usepackage{tabularx}
\usepackage{multirow}
\usepackage{physics}
\usepackage{siunitx} % for \degree
\usepackage{listings}
\usepackage{microtype}
\usepackage[most]{tcolorbox}
\tcbset{
  colback=gray!5,        % background color
  colframe=gray!60,      % border color
  fonttitle=\bfseries,   % bold title
  coltitle=white,        % title text color
  title filled=true,
  sharp corners,
  boxrule=0.4pt,
}

\newtcolorbox{llmprompt}[1][]{
  colback=gray!10,
  colframe=black!70,
  boxrule=0.5pt,
  arc=1pt,
  left=6pt,
  right=6pt,
  top=6pt,
  bottom=6pt,
  fonttitle=\bfseries,
  title=#1
}

\newtcolorbox{llmpromptbox}[1][]{
  colback=gray!5,
  colframe=gray!60,
  fonttitle=\bfseries,
  coltitle=white,
  title filled,
  title=LLM prompt,
  #1
}

\newtcolorbox{llmpromptboxsmall}[1][]{
  colback=gray!5,
  colframe=gray!60,
  fonttitle=\small\bfseries,
  coltitle=white,
  title filled,
  title=LLM prompt,
  #1
}

\newtcolorbox{promptbox}{title=\textbf{Prompt}, colback=blue!3}
\newtcolorbox{toolbox}{title=\textbf{Tool use (queries \& calls)}, colback=green!5}
\newtcolorbox{evidencebox}{title=\textbf{Retrieved evidence}, colback=orange!5}
\newtcolorbox{decisionbox}{title=\textbf{Controller decision}, colback=purple!5}
\newtcolorbox{rationalebox}{title=\textbf{Brief rationale}, colback=purple!5}

\lstdefinestyle{sqlstyle}{
  basicstyle=\ttfamily\footnotesize,
  breaklines=true,
  breakatwhitespace=false,
  columns=fullflexible,
  keepspaces=true,
  frame=none
}

\journal{Elsevier}
\begin{document}

\begin{frontmatter}
    \title{Evaluating LLMs as Interpretable Controllers for Dynamical Systems}
    \author[label1]{Aleksander Østensen}
    \author[label1]{Alberto Mino Calero}
    \author[label1]{Anastasios M. Lekkas}
    \author[label1,label2]{Adil Rasheed\corref{mycorrespondingauthor}}    
    \cortext[mycorrespondingauthor]{Corresponding author}

    \ead{adil.rasheed@ntnu.no}
    \affiliation[label1]{
    organization={Department of Engineering Cybernetics, NTNU},
            addressline={O.S. Bragstads plass 2E}, 
            city={Trondheim},
            postcode={7034}, 
            %state={},
            country={Norway}
            }
    \affiliation[label2]{
    organization={Mathematics and Cybernetics, SINTEF Digital},
            addressline={Sem Sælands vei 11}, 
            city={Trondheim},
            postcode={7034}, 
            %state={},
            country={Norway}
            }
            
\begin{abstract}
Large Language Models (LLMs) are increasingly used for decision-making and reasoning tasks, yet their potential as controllers for physical systems remains largely unexplored. This work investigates whether LLMs can function as interpretable controllers for a dynamic thermal environment, examining their ability to follow setpoints, interpret natural-language commands, reason about actuator effects, and incorporate prior model-based knowledge. Five LLMs of varying scales are evaluated under multiple scenarios, including settings with penalties on heater or fan usage and cases where the models have access to a physics-based prediction tool. The results show that control performance depends on model complexity: while low- and mid-scale models frequently misinterpret actuator dynamics or generate inconsistent reasoning, high-complexity models such as Qwen-3~14B and GPT-4o achieve accurate temperature tracking, stable actuator usage, and coherent explanations aligned with physical principles. Incorporating a physics-based model significantly improves control smoothness and energy efficiency by enabling anticipatory decision-making. A detailed reasoning taxonomy further reveals a clear progression from causal misinterpretation in smaller models to cohesive and temporally aware reasoning in larger ones. The findings demonstrate that LLMs can act as interpretable controllers when sufficiently capable and appropriately grounded in domain knowledge, highlighting promising opportunities for hybrid model-based and language-driven control strategies that can provide plausible explanations. 
\end{abstract}

\end{frontmatter}

\section{Introduction}
\label{sec:introduction}
Control of dynamical systems traditionally relies on mathematically grounded methods such as optimal control, reinforcement learning, and model predictive control. These approaches provide strong guarantees on stability and performance when accurate system models are available. However, they often struggle with systems that are difficult to model, tasks requiring high-level reasoning, or scenarios where the interpretability of control actions is essential. Recent advances in Large Language Models (LLMs) have shown remarkable capabilities in reasoning, planning, and interacting with structured and unstructured information, skills that open new possibilities for their use beyond language understanding.

LLMs have emerged as powerful sequence learners capable of understanding, generating, and transforming natural language with remarkable fluency and awareness of the context. Their ability to follow instructions, display reasoning capabilities, and incorporate diverse modalities has made them central components in modern Artificial Intelligence (AI). Transformer-based architectures \cite{Vaswani2023aia} enable these models to capture long-range dependencies and semantic relationships, which in turn allow them to reliably interpret natural-language descriptions of system states, objectives, and constraints, an ability directly relevant to natural-language-guided control.

LLMs have demonstrated strong performance across a wide range of application domains. In robotics, they have been used for task planning, action sequencing, and grounding natural-language instructions into executable actions \cite{Ichter2023dai,Zitkovich2023r2v,Huang2023ime}. In software engineering, LLMs assist in code generation \cite{Chen2021ell}, and tool-use automation \cite{Schick2023tlm}. In scientific domains, they have been applied to materials discovery \cite{Pyzer_Knapp2025fmf} and biomedical information extraction \cite{Singhal2023llm}. These successes highlight the adaptability of LLMs and their capacity to operate over symbolic structures, domain knowledge, and natural-language descriptions of complex tasks.

Recently, LLMs have begun to attract interest as components in decision-making and control pipelines. Several studies have explored LLM-based planning or policy synthesis for embodied agents \cite{Huang2022lma,Song2023lpf}, showing that LLMs can transform high-level natural-language goals into actionable plans. Extensions toward reinforcement learning have examined how LLMs can critique profiles \cite{Stiennon2020lts}, guide policy search \cite{Hu2023lir}, or serve as world-model priors \cite{Lin2024ltm}. More closely related to control, LLMs have been used to interpret system dynamics in open-loop reasoning \cite{Wang2025llm}, or provide human-readable justifications for selected actions. These studies suggest that LLMs possess emergent capabilities such as commonsense reasoning, symbolic manipulation, and natural-language explanation that could make them promising candidates for interpretable control frameworks.

More recently, LLMs' potential as a closed-loop controller was evaluated in \cite{Rasheed2025llm} and \cite{Rasheed2025hms}, where it was found to be at par with both MPC and RL, with the advantage that not much knowledge is needed to build such controllers. Despite these advances, a systematic evaluation of LLMs as closed-loop controllers for dynamical systems remains limited, and even more so from an interpretability perspective. In particular, questions persist regarding their reliability, interpretability, and capacity to leverage prior domain knowledge during real-time action selection. This motivates the present study, which investigates these aspects across a range of LLM scales and modeling assumptions.

LLMs possess several features that make them intriguing candidates for control applications. First, they can integrate diverse forms of knowledge (mathematical descriptions, symbolic relations, natural language instructions, and demonstrations), enabling a form of model-based reasoning without explicit parameterization. Second, they can produce chain-of-thought explanations, offering transparency and interpretability rarely available in classical control or deep reinforcement learning systems. Although their reasoning traces are still part of the model's output and may not be faithful to the underlying process followed to reach the output \cite{Turpin2023lmd}, the coherence and plausibility of the chain-of-thought explanations, actions, and system states can be analyzed together to assess their understanding of the system being controlled. Finally, LLMs exhibit strong generalization across tasks and domains, suggesting that they may learn control strategies for systems even when training data or models are incomplete, as long as their reasoning capabilities and understanding of the system and actions are sound.

Existing work on LLM-based decision making, such as LLMs in robotics planning, tool use, and reasoning-based reinforcement learning, indicates that language models can approximate policies, evaluate system states described in natural language, and construct action sequences using symbolic knowledge. However, their use as controllers for continuous or hybrid dynamical systems remains largely unexplored. In particular, questions remain about whether LLMs can reliably generate control actions, how much they depend on prior model knowledge, and whether their decisions can be trusted or interpreted in safety-critical settings.

Motivated by these gaps, this work investigates the feasibility of employing LLMs as interpretable controllers. We aim to understand not only their control capabilities, but also the mechanisms underlying their decisions and the conditions under which they succeed or fail. Specifically, we address the following research questions:

\begin{itemize}
    \item Can LLMs effectively control dynamical systems?
    \item Can LLM make sense of the commands issued in natural language? % This one is already known to be true, since they can follow instructions.
    \item Does incorporating prior model-based knowledge improve their control performance?
    \item Can LLMs generate meaningful explanations for their control decisions?
    \item How does the quality of reasoning and explanation vary across LLMs of different scales?
    % \item How does uncertainty in the provided prior knowledge affect LLM-generated control recommendations?
\end{itemize}
To address these questions, we evaluate five LLMs spanning a wide range of scales and architectures. The models are tested both with and without access to a physics-based predictive tool, under scenarios that introduce control penalties. This enables a systematic analysis of closed-loop control performance, energy efficiency, and the physical plausibility of LLM-generated explanations. Through these experiments, we aim to assess not only how well LLMs can control a real-world system, but also whether their decisions can be trusted, interpreted, and improved through the incorporation of domain knowledge.

The remainder of this article is structured as follows. Section~\ref{sec:theory} reviews the theoretical foundations of LLMs and the physics-based model used in the study. Section~\ref{sec:method} describes the experimental setup, prompt design, and controller framework. Section~\ref{sec:resultsanddiscussions} presents the results and discusses the control performance and reasoning capabilities of the evaluated models. Section~\ref{sec:conclusionandfuturework} concludes the paper by summarizing the key findings and outlining directions for future work.

\section{Theory}
\label{sec:theory}

\subsection{Large Language Models}
\label{LMMs}

Large language models (LLMs) are transformer-based models trained on large corpora of text and code using next-token prediction objectives. The inductive bias of the models gained through their training, instruction-tuning, and safety alignment enables the emergence of general-purpose abilities, such as reasoning, planning, following instructions, and providing information. 

In this work, we test their ability to be used as black-box controllers that map natural-language prompts describing the system state and goals into structured control actions. Our focus is to test the capacity of open, local LLMs of different complexities, representatives of different sizes, training regimes, and general capabilities. For this, we selected 4 models Gemma-3 \cite{team_gemma_2025} with 1B parameters, Phi-4 Mini \cite{microsoft_phi-4-mini_2025} with 3.8B, Mistral \cite{jiang_mistral_2023} with 7B, and Qwen-3 \cite{yang_qwen3_2025} with 14B. Additionally, we also test GPT-4o \cite{OpenAI2024g4t1} to improve the comparison with one of the most complex LLMs available. A comparison summary can be found in Table~\ref{tab:LLM_summary}.

\paragraph{Gemma-3 1B}
Gemma-3 1B \cite{team_gemma_2025} is a compact multimodal model with 1B parameters, also available in other sizes, that is designed specifically for local deployment. Its architecture has grouped-query attention (GQA) and interleaved local and global attention layers that allow long context lengths, specifically 32K for the 1B variant used in this work. It was trained on 2 trillion tokens with data that includes web documents in multiple languages, code, mathematics, and images, and can follow instructions and provide basic reasoning, understanding both images and text. The model has a vocabulary of 256K tokens, and also features a knowledge-distillation training \cite{hinton_distilling_2015}, where Gemma-3 was learned from a higher-complexity teacher model. In our setting, Gemma-3 provides a representative example of low-capacity models capable of following instructions, which opens the possibility of also showing controlling capabilities. 

\paragraph{Phi-4 Mini 3.8B}
Phi-4 Mini \cite{microsoft_phi-4-mini_2025} is another low-complexity model, with only 3.8B parameters, that was trained with a corpus that emphasizes mathematical reasoning, programming, and following instructions. Its architecture contains GQA and supports a longer context length of 128K tokens, allowing the model to process longer sequences than Gemma-3 1B. One of its particularities is that its training relies heavily on synthesized data, in addition to the web documents contained in the corpus, which amounts to a total size of 5 trillion tokens for training. Also, the training data is more targeted, rather than large-scale web scraping, and this provides the model with more efficient reasoning capabilities in structured reasoning tasks for its parameter count. Although Phi-4 also has a multimodal version, in this work, we only test the Mini version, which understands text but not other types of data, which has a vocabulary of 200K tokens. In our study, this model represents an LLM trained for efficiency using higher-quality data, and it provides a reference to evaluate whether the enhanced quality data and larger size can truly make a difference in control-oriented reasoning.

\paragraph{Mistral 7B}
Mistral \cite{jiang_mistral_2023} is a slightly higher-complexity LLM with 7B parameters that also features GQA and sliding-window attention in its architecture with the aim of improving the efficiency with which the model handles long contexts, allowing the model to use a lower context length of only 8K. Mistral was likely trained on large-scale text corpora, although the amount or source of the data used to train the model is not disclosed, and has a vocabulary of 32K tokens. The model focuses on good performance following instructions relative to its size, prioritizing inference speed and memory efficiency while keeping robust general-purpose reasoning capabilities. Mistral 7B serves as a mid-complexity model in terms of parameters, while featuring lower context lengths and vocabulary than Gemma-3 and Phi-4 Mini, offering insight into how these different architectures balance the instruction following capabilities with understanding state description and keeping stable behaviors in the multi-step closed-loop control.

\paragraph{Qwen-3 14B}
Qwen-3 14B \cite{yang_qwen3_2025} is a dense transformer of high complexity, compared with the previous ones, that benefits from GQA while featuring query/key normalization (QK-Norm) to increase the stability of the training process. The model supports a context length of 128K tokens, intermediate compared with the other models tested in this work. It was trained with a corpus with 119 languages in diverse domains such as coding, STEM (Science, Technology, Engineering, and Mathematics), reasoning tasks, books, and synthetic data, for a total of 36 trillion tokens, whilst having a vocabulary of 151K tokens. Qwen-3 has the particularity that, although it was trained to follow instructions like the others, it also focuses on following a chain-of-thought reasoning strategy. It was also fine-tuned across a wider range of reasoning and tool-use tasks, showing ability as an AI agent, making it more suitable for applications that require structured decision-making. Qwen-3 represents a high-complexity model with strong general-purpose reasoning capabilities, which makes it a good fit for stable control behaviors with enough depth to anticipate changes in the system, which should enhance its control strategy.

\paragraph{GPT-4o}
GPT-4o \cite{OpenAI2024g4t1} is the only proprietary model used in our experiments and, although its exact number of parameters is unknown, it is likely to be the one with the highest complexity. Being a closed model, specific details about its architecture and training are not fully disclosed, but it is known to be trained on next-token prediction followed by post-training alignment, similarly to most contemporary LLMs. It is also multimodal, understanding both text and images, and exhibits the strongest reasoning, instructions following, and safety-aligned behavior according to benchmarks. However, it is known that its context length is 128K tokens and has a vocabulary of almost 200K tokens \cite{yang_problematic_2024}. GPT-4o has demonstrated human-level and reliable performance on a wide range of professional and academic benchmarks, providing our experiments with a state-of-the-art high-complexity model with outstanding reasoning capabilities and, possibly, the best-fitting control strategies in our setting. 

\begin{table*}[!htbp]
    \centering
    \caption{LLMs comparison summary.}
    \label{tab:LLM_summary}
    \resizebox{\textwidth}{!}{
        \begin{tabular}{lcccccc}
        \toprule
        \textbf{Model} & \textbf{Parameters} & \textbf{Complexity} & \textbf{Key features} &
        \textbf{Context} & \textbf{Pre-training size} & \textbf{Vocabulary}\\
        \midrule
        Gemma-3     & 1B        & Low     & GQA, local/global attn, distillation, multimodal    & 32K   & 2T  & 256K \\
        Phi-4 Mini  & 3.8B      & Low     & GQA, curated \& synthetic data    & 128K  & 5T  & 200K \\
        Mistral     & 7B        & Mid     & GQA, sliding-window attention    & 8K    & Unknown  & 32K \\
        Qwen-3      & 14B       & High    & GQA, QK-Norm, multilingual, STEM    & 128K  & 36T  & 151K \\
        GPT-4o      & Unknown   & High    & Multimodal, alignment post-training    & 128K  & Unkonwn & 200K \\
        \bottomrule
        \end{tabular}
    }
\end{table*}

\subsection{Physics-Based Modeling}
\label{subsubsec:theorypbm}
The dynamical system used in this study is a cuboidal box measuring $50cm \times 50cm \times 60cm$. The inside temperature is controlled using a heater and high-flow intake and outtake fans. Based on the principles of energy conservation, the PBM is given by the following equation:
\begin{equation}
    \frac{dT}{dt} = \frac{H}{\rho V C_p} - \frac{F (T - T_{amb})}{V}
\label{eq:pbm}
\end{equation}
where the term on the left-hand side represents the change in inside temperature with respect to time, while on the right-hand side, the first term represents the heat input from the heater, where $H$ is the heating power, $\rho$ is the air density, $V$ is the enclosure volume, and $C_p$ is the specific heat capacity of air. Since the heater is operated using a duty cycle $u_h$, $H$ is expressed as $u_{h}H_{max} $. The $u_h$ can be changed in discrete steps of 0.05 from 0 to 1. The second term represents the heat exchange induced by the fan, where the airflow is modeled as $F = u_{f}F_{max}$, with $u_f$ denoting the fan’s ON/OFF state. $T$ is the internal air temperature, and $T_{amb}$ is the external ambient temperature. 

This model is based on several assumptions. The air inside the enclosure is considered to be perfectly mixed, resulting in a uniform temperature distribution. The properties $\rho$ and $C_p$ are assumed to be constant. The heater and fan are assumed to respond instantaneously, delivering $H$ amount of heat and moving $F$ volume of air per second, respectively. Furthermore, it is assumed that no air flows through the fan when it is off and that no heat is lost through the enclosure walls. The effects of plants on temperature and flow are also ignored. 

\section{Method}
\label{sec:method}

\subsection{System components}
\begin{figure}[!htbp]
    \centering
    \includegraphics[width=\linewidth,trim= 20 10.5 20 12,clip]{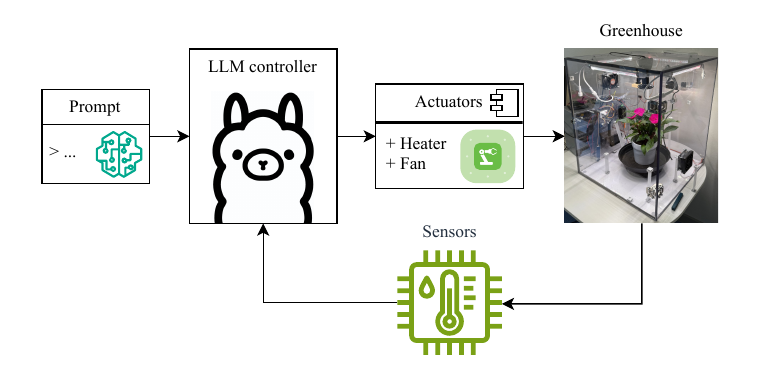}
    \caption{Schematic.}
    \label{fig:schematic}
\end{figure}

%The LLM controller consists of two agents; the planner agent and the critique agent. The role of the planner agent is to generate control plans, while the critique agent gives feedback on how well the plans reduce error.
The LLM controller is set up as a langgraph agent, named planner. The planner agent was configured with LangChain's ChatOllama class. This treats the model as a chat agent that receives messages throughout the workflow. This message feature has been used to provide the model with continuous updates of values from the simulated model, telling it if its output plan is invalid or if a tool call failed. A general diagram of the components of the system can be found in Figure~\ref{fig:schematic}.

\begin{figure}[t]
  \centering
  \begin{subfigure}[t]{0.48\linewidth}
    \centering
    \includegraphics[width=\linewidth]{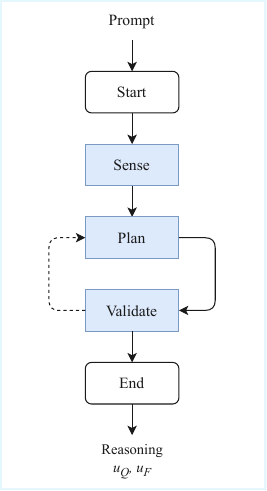}
    \caption{Simple model.} 
    \label{fig:model-left}
  \end{subfigure}\hfill
  \begin{subfigure}[t]{0.48\linewidth}
    \centering
    \includegraphics[width=\linewidth]{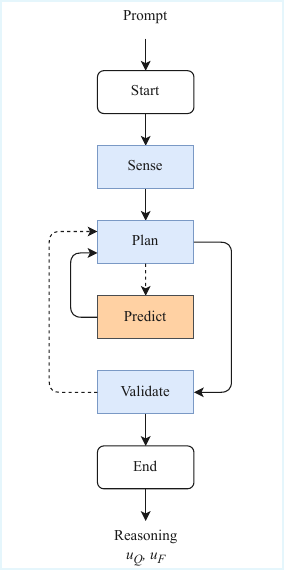}
    \caption{Prediction model.}
    \label{fig:model-right}
  \end{subfigure}

  \caption{LLM controller workflows. The blue boxes are nodes that the controller will always enter in its runtime. The orange box symbolizes a tool that the model can call upon. The dashed arrows are conditional edges. If a plan has been deemed invalid by the validation node, the model will be asked to provide a new plan. If the output from the model includes a tool call, the model will enter the prediction node and get the predicted results for a given candidate plan.}
  \label{fig:pipelines-two}
\end{figure}

The ChatOllama class also supports the use of structured outputs, forcing the model to deliver its response in JASON format. This approach provides a consistent form of output from the LLM, which in turn can automatically be extracted and turned into actionable control inputs without relying on fragile text parsing. The workflow for the models with and without the prediction tool is shown in Figure~\ref{fig:pipelines-two}.

\subsection{Experiments}
The experimental evaluation is organized around three axes: control capability, sensitivity to multi-objective constraints, and the effect of domain-knowledge grounding. For the first axis, all five LLMs are tested on their ability to track a time-varying reference temperature profile over approximately 360 minutes using 60-second control steps, with no restrictions on actuator usage. The second axis introduces penalty scenarios that prioritize minimizing the usage of a specific actuator, either the heater or the fan, while still requiring accurate temperature tracking. These penalties are communicated to the models exclusively through modifications to the natural-language prompt, testing whether LLMs can interpret and balance competing objectives expressed in plain language. The third axis investigates whether providing the two best-performing models (Qwen-3 14B and GPT-4o) with access to a physics-based prediction tool improves their control behavior. 

In the tool-assisted setting, the models can call a forward simulator to evaluate candidate action sequences over a 10-minute planning horizon before committing to a control action. Across all scenarios, the models operate in closed loop: at each timestep, the LLM receives the current temperature, the ambient temperature, and the setpoint, and must return a heater duty cycle and a binary fan command together with a natural-language explanation of its reasoning. This explanation is subsequently analyzed to assess the interpretability and physical coherence of each model's decision-making process.

\subsection{Tuning of PBM}
The physics-based model is based on Equation~\ref{eq:pbm}. When the model is used to predict the temperature, the solver RK45 from the Python Library SciPy \cite{Virtanen2020s1f} is used to solve the ODE. The time delta of $\Delta t = 60s$ is used as the maximum integration time for each timestep. Initially, the predicted values were not close to the measured temperature. This is because the PBM assumes a uniform temperature inside the volume. However, in reality, the temperature has a spatial variation, with the highest temperature around the heater and then gradually receding away from it. Due to this, the rated values for the heater and the fan, $H_{\text{max}}$ and $F_{\text{max}}$, were not used. Instead, these values were estimated by finding the mean of the measured values. For $H$, this was done by finding two consecutive data points in the dataset where the heater duty cycle was set to 1, and the fan was off. For each of these data points, the average measured heat exchange $\Bar{H}_{\text{max},t}$ was calculated using Equation~\ref{eq:method-max-q}. The new value $H^{'}_{\text{max}}$ was found by averaging the measured values of $\Bar{H}_{\text{max},t}$. 

\begin{equation}
    \Bar{H}_{\text{max},t} = \rho V C_p \frac{(T_{t} - T_{t-1})}{dt}
\label{eq:method-max-q}
\end{equation}

For the fan, the dataset was filtered to include data points where the heater duty cycle was 0 and the fan was on. The measured average air flow $\Bar{F}_{\text{max},t}$ was found using Equation~\ref{eq:method-max-x}. 

\begin{equation}
    \Bar{F}_{\text{max},t} = - V \frac{(T_{t} - T_{t-1})}{(T_{t-1} - T_{\text{ambient}})dt}
\label{eq:method-max-x}
\end{equation}

The new value $F^{'}_{\text{max}}$ was determined by averaging the recorded measurements of $\Bar{F}_{\text{max},t}$, and these new values resulted in improved model performance.

\section{Results and discussions}
\label{sec:resultsanddiscussions}
This section presents and discusses the performance of the evaluated LLM-based controllers across the different experimental scenarios. The analysis is organized around three main aspects: the ability of the models to regulate the temperature in a closed loop, their response to actuator-usage penalties, and the effect of providing prior domain knowledge through a physics-based predictive model. In addition to quantitative control performance, the section also examines the explanations produced by the models in order to assess whether their control decisions are physically meaningful, internally consistent, and aligned with the observed system dynamics. Together, these results provide insight into both the practical control capability of LLMs and their potential role as interpretable controllers for dynamical systems.

\subsection{Can LLMs control a dynamical system ?}
The first set of experiments evaluates whether LLMs can act as closed-loop controllers for the dynamical system under consideration. In these experiments, each model receives the current temperature, ambient temperature, and reference setpoint, and must return suitable control actions for the heater and fan. The objective is to assess whether the models can translate natural-language descriptions of the control task into physically appropriate actuator commands. The evaluation is carried out under three conditions: without any actuator penalty, with a penalty on heater usage, and with a penalty on fan usage. This makes it possible to compare not only the basic tracking capability of the models, but also their ability to handle competing objectives expressed through natural language.

\begin{figure*}[!htbp]
  \centering
  % LEFT column
  \begin{minipage}[t]{0.48\textwidth}
    % \centering
    % Top: temperature trajectories
    \begin{subfigure}{\linewidth}
    \centering
    \includegraphics[width=\linewidth]{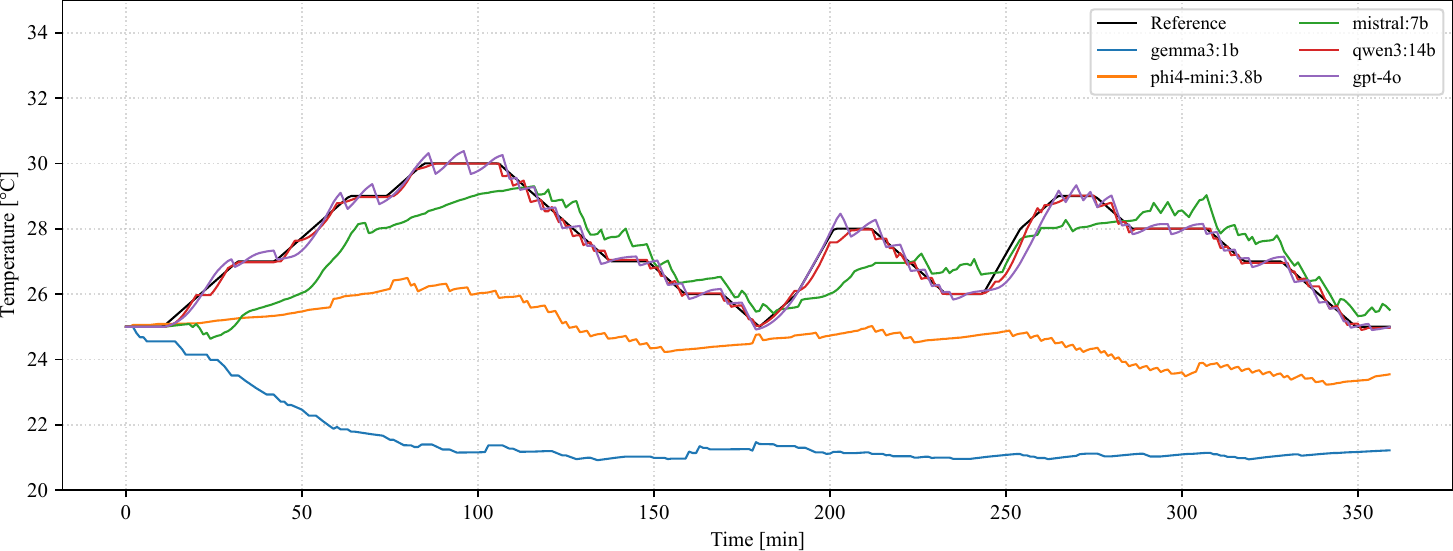}
    \caption{Temperature profile.}
    \label{fig:profile2-no-penalty-temp-overlay}
    \vspace{0.2cm}
    \end{subfigure}
    % Bottom-left: fan sequences
    \begin{subfigure}[t]{0.49\linewidth}
    \centering
    \includegraphics[width=\linewidth]{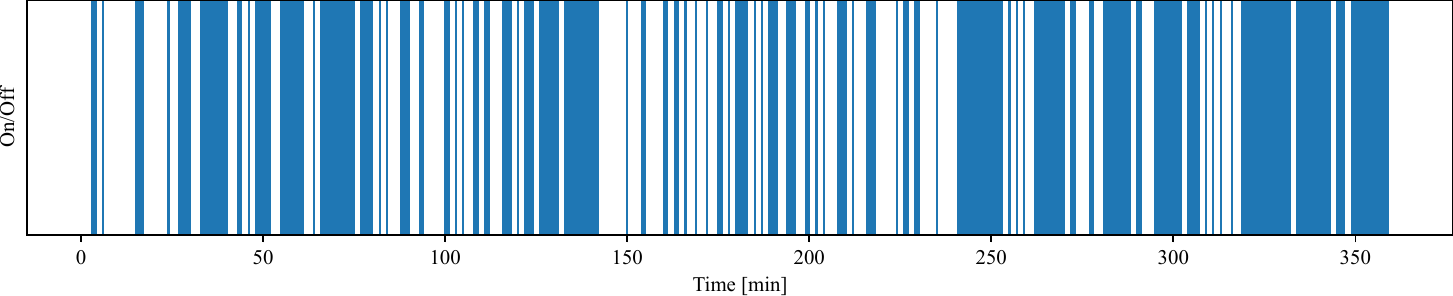}\par\vspace{0.35em}
    \includegraphics[width=\linewidth]{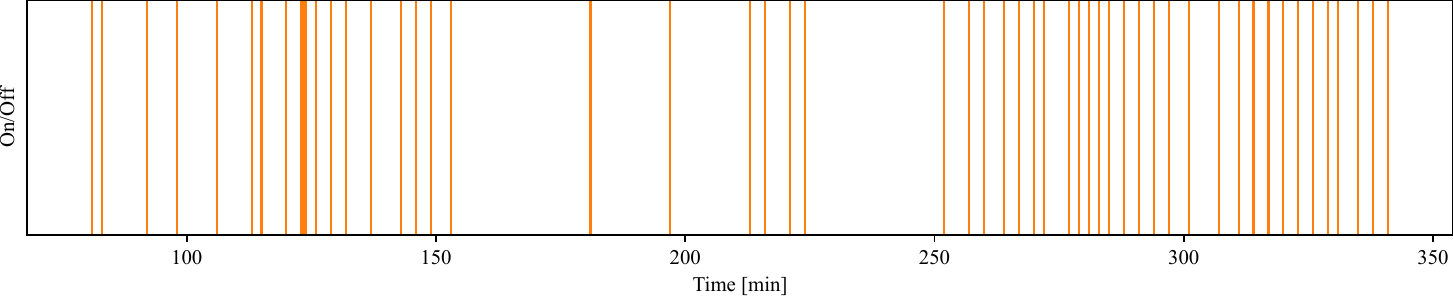}\par\vspace{0.35em}
    \includegraphics[width=\linewidth]{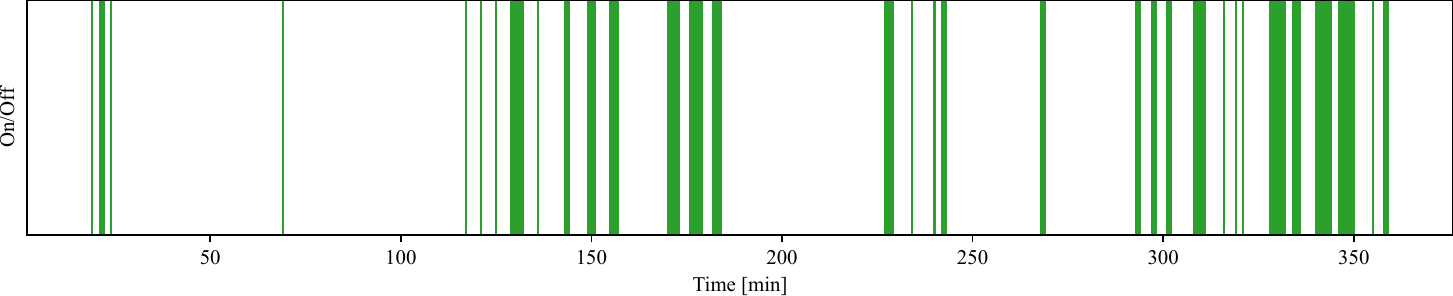}\par\vspace{0.35em}
    \includegraphics[width=\linewidth]{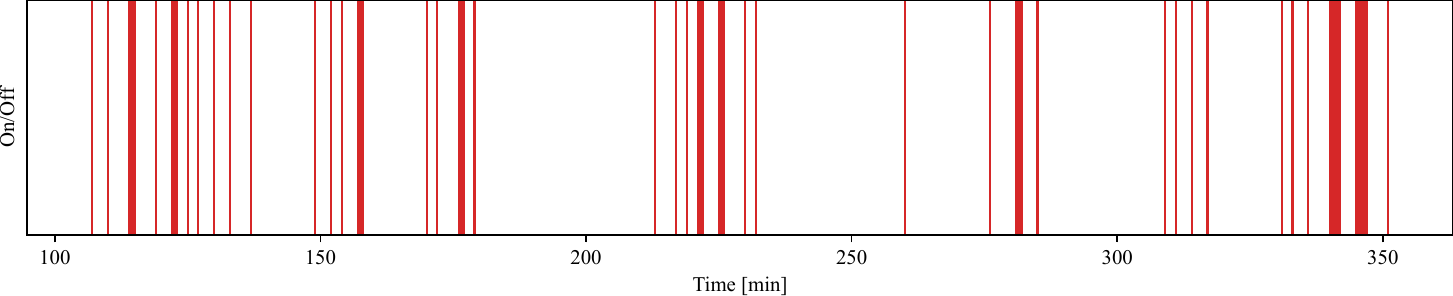}\par\vspace{0.35em}
    \includegraphics[width=\linewidth]{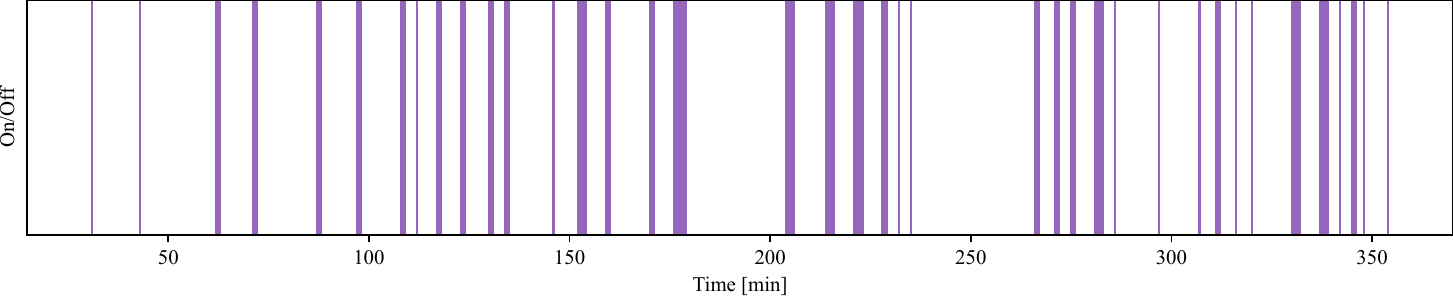}
    \caption{Fan sequences.}
    \label{fig:profile2-no-penalty-fan}
    \end{subfigure}
    \hfill
    % Bottom-right: heater duty cycle
    \begin{subfigure}[t]{0.49\linewidth}
    \centering
    \includegraphics[width=\linewidth]{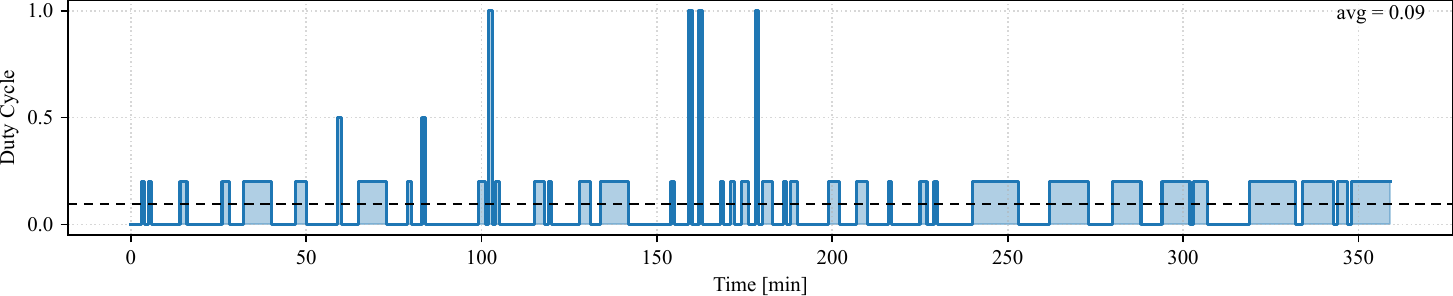}\par\vspace{0.35em}
    \includegraphics[width=\linewidth]{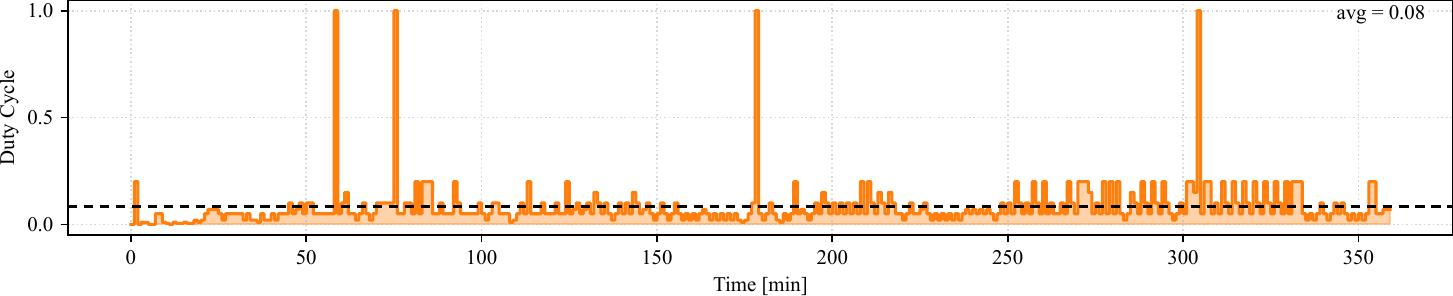}\par\vspace{0.35em}
    \includegraphics[width=\linewidth]{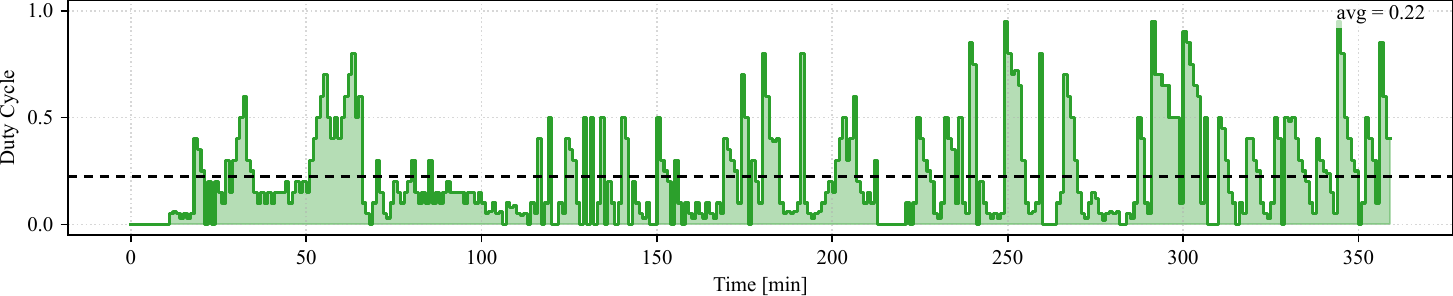}\par\vspace{0.35em}
    \includegraphics[width=\linewidth]{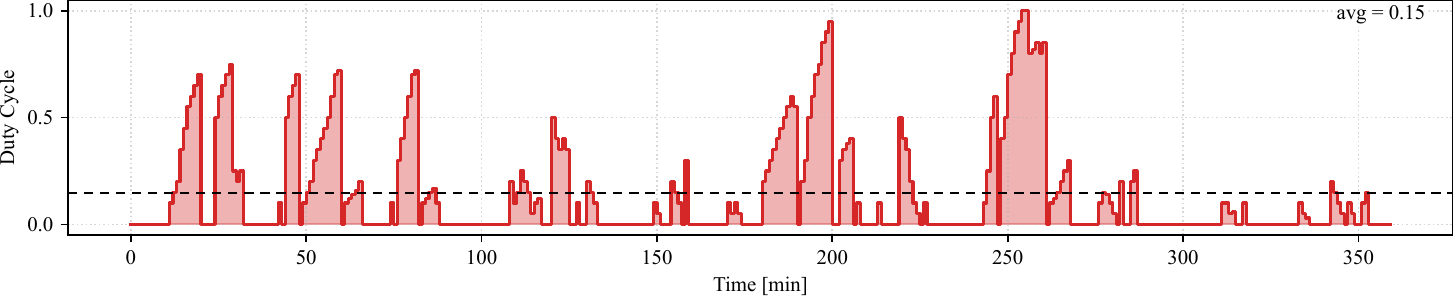}\par\vspace{0.35em}
    \includegraphics[width=\linewidth]{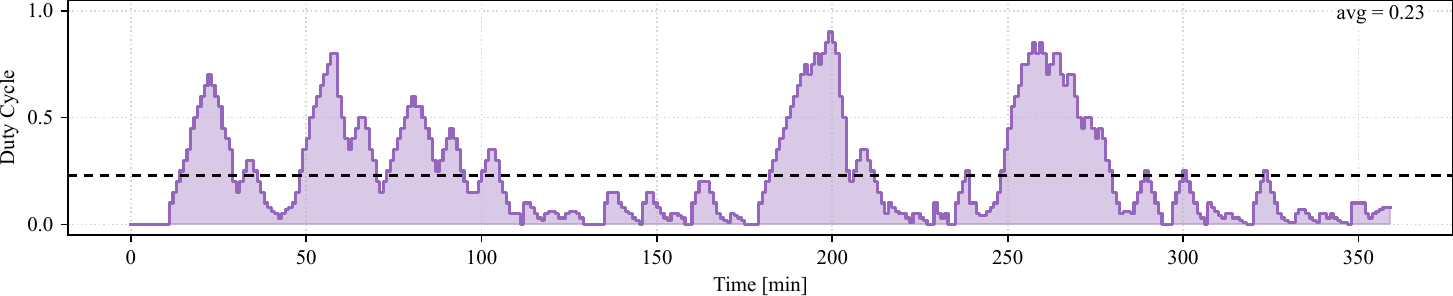}
    \caption{Heater duty cycle.}
    \label{fig:profile2-no-penalty-heater}
    \end{subfigure}
    \caption{Temperature profile and control sequences.}
    \label{fig:profile2-no-penalty-overview}
    \vspace{0.54cm}
    % Top: temperature trajectories
    \begin{subfigure}{\linewidth}
    \centering
    \includegraphics[width=\linewidth]{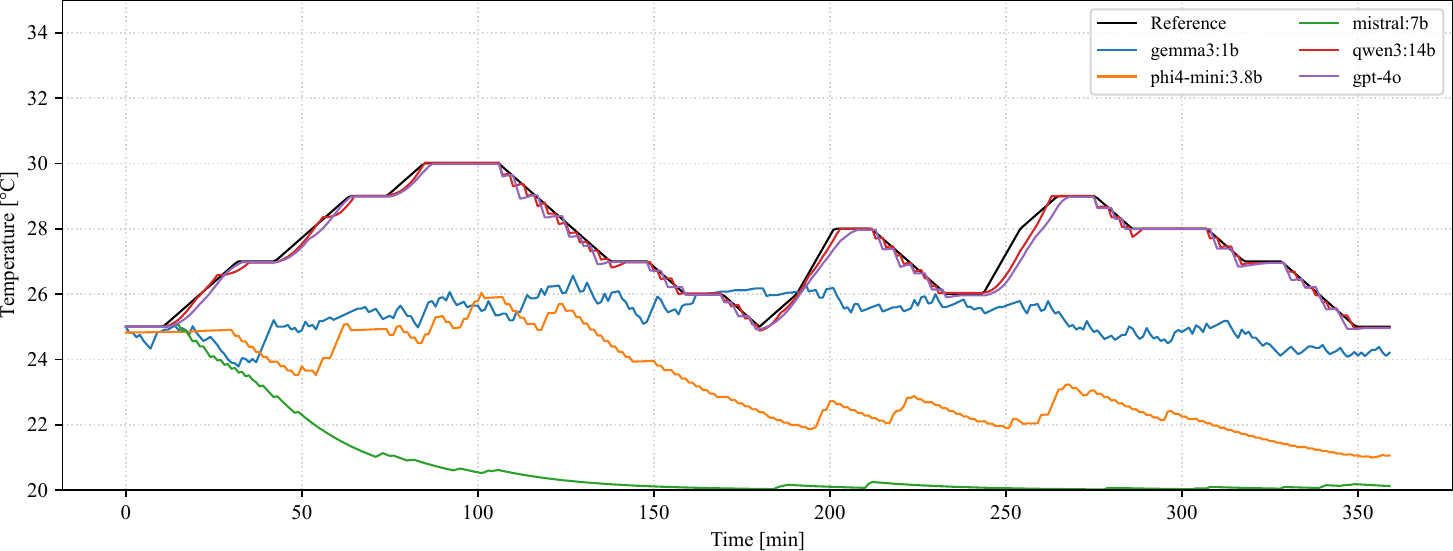}
    \caption{Temperature profile with heater penalty.}
    \label{fig:profile2-heater-penalty-temp-overlay}
    \vspace{0.2cm}
    \end{subfigure}
    % Bottom-left: fan sequences
    \begin{subfigure}[t]{0.49\linewidth}
    \centering
    \includegraphics[width=\linewidth]{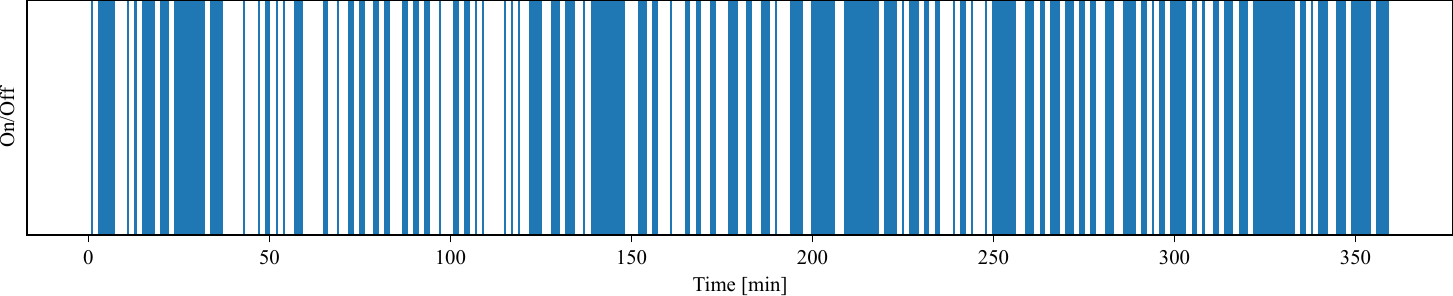}\par\vspace{0.35em}
    \includegraphics[width=\linewidth]{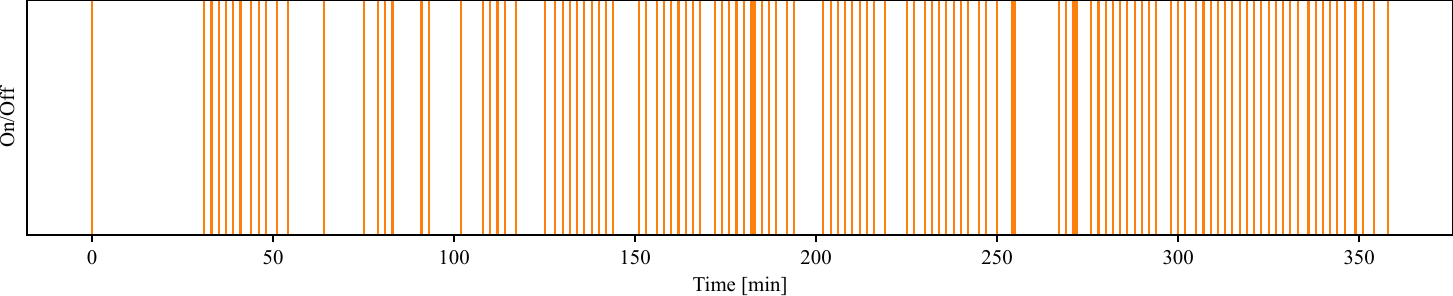}\par\vspace{0.35em}
    \includegraphics[width=\linewidth]{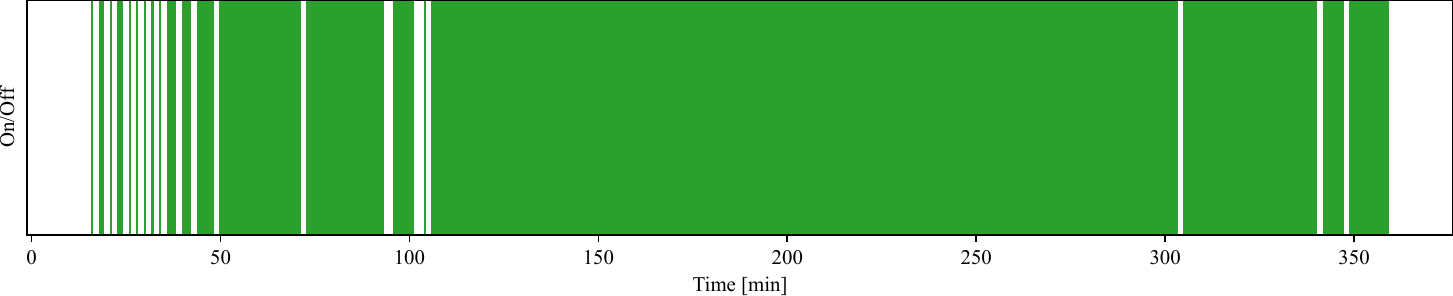}\par\vspace{0.35em}
    \includegraphics[width=\linewidth]{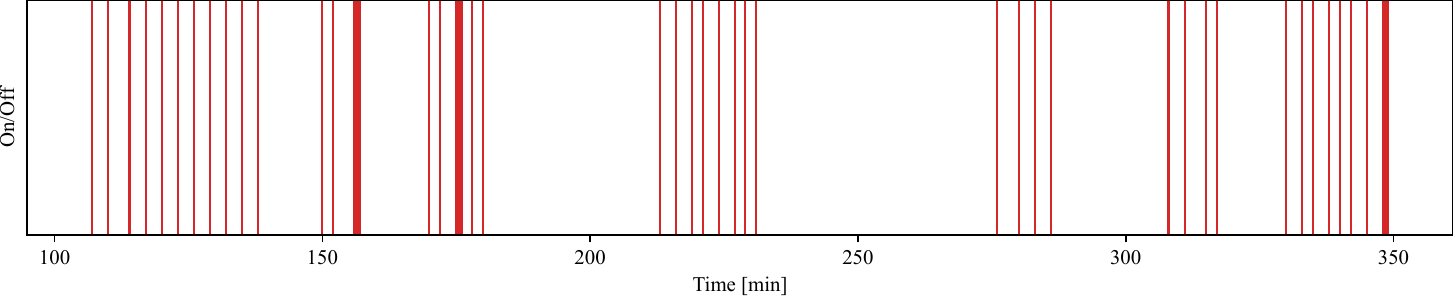}\par\vspace{0.35em}
    \includegraphics[width=\linewidth]{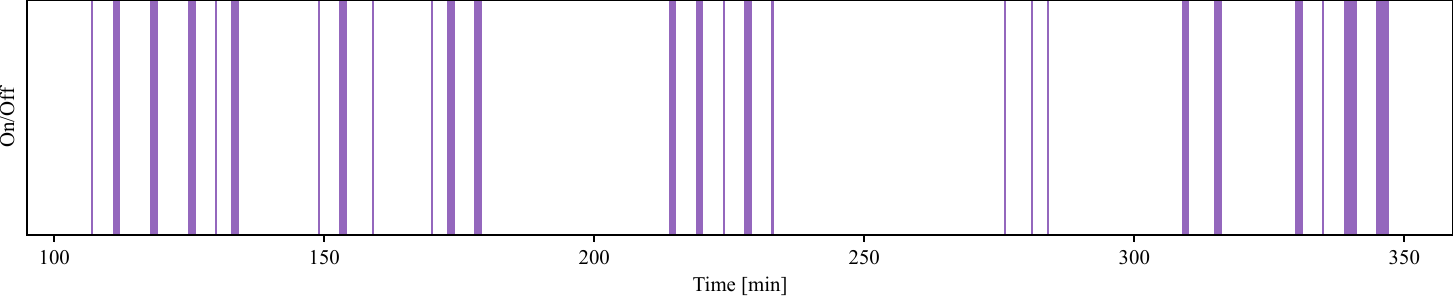}
    \caption{fan sequences with penalty.}
    \label{fig:profile2-heater-penalty-fan}
    \end{subfigure}
    \hfill
    % Bottom-right: heater duty cycle
    \begin{subfigure}[t]{0.49\linewidth}
    \centering
    \includegraphics[width=\linewidth]{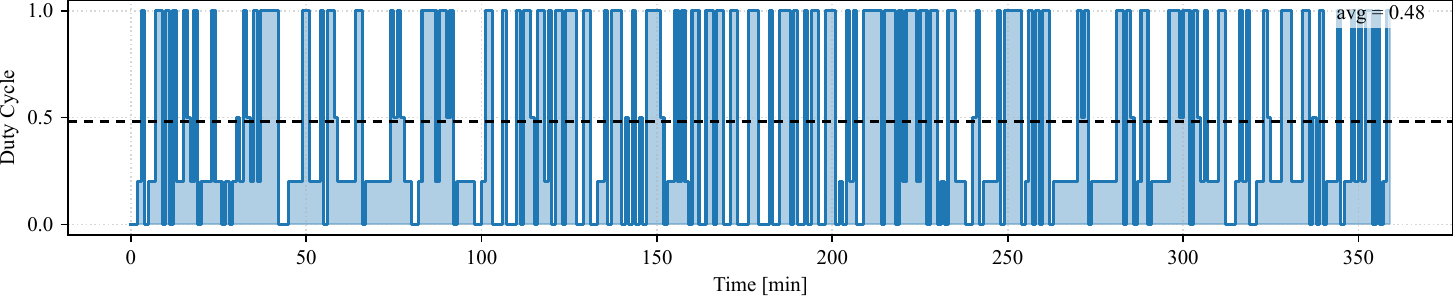}\par\vspace{0.35em}
    \includegraphics[width=\linewidth]{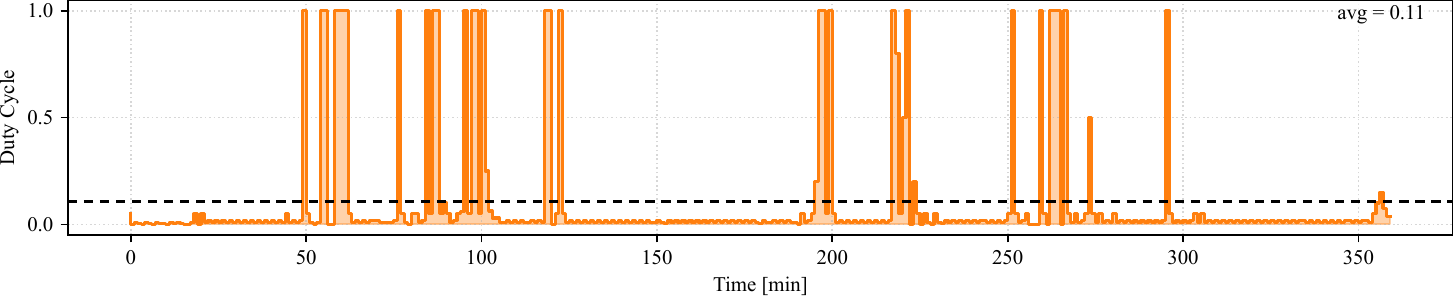}\par\vspace{0.35em}
    \includegraphics[width=\linewidth]{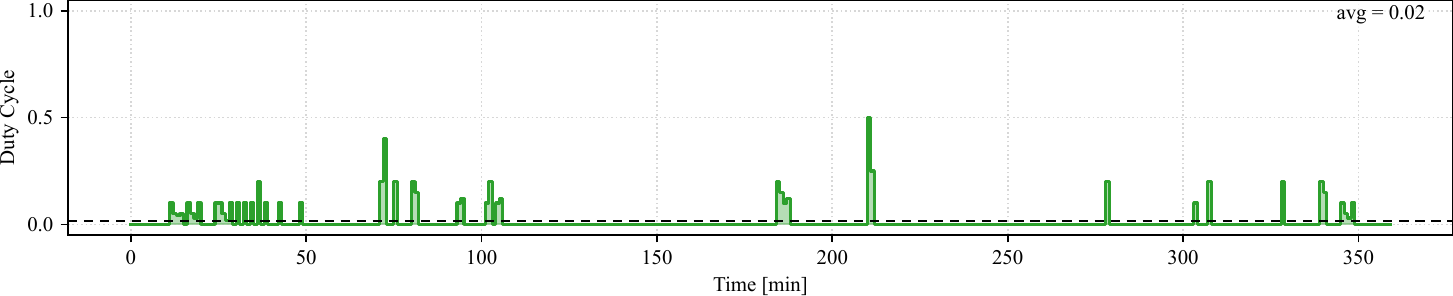}\par\vspace{0.35em}
    \includegraphics[width=\linewidth]{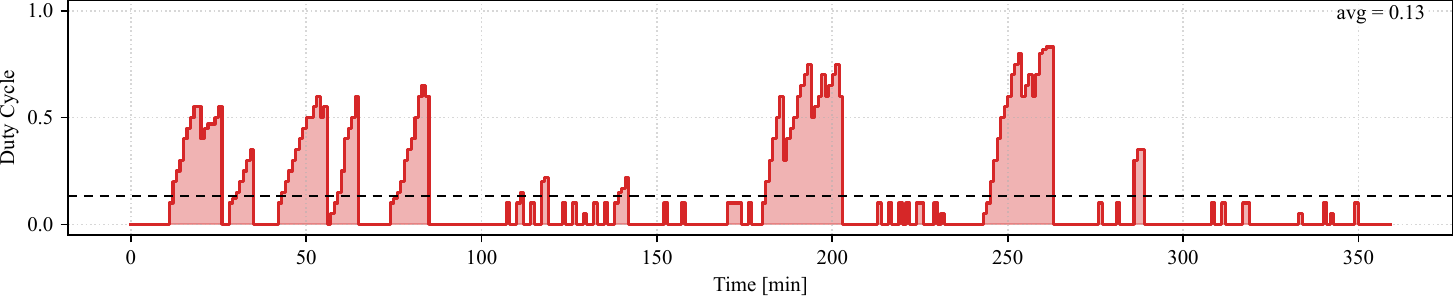}\par\vspace{0.35em}
    \includegraphics[width=\linewidth]{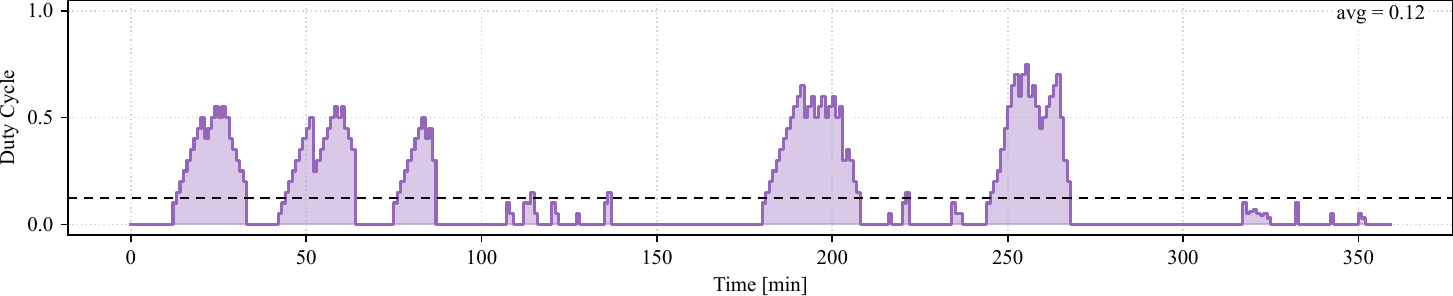}
    \caption{Heater duty cycle with penalty.}
    \label{fig:profile2-heater-penalty-heater}
    \end{subfigure}
    \caption{Temperature profile and control sequences with heater penalty.}
    \label{fig:profile2-heater-penalty-overview}
  \end{minipage}%
  \hfill
  % RIGHT column
  \begin{minipage}[t]{0.48\textwidth}
    \vspace{-25.48ex} %Needs to be adjusted manually to avoid empty space before the figure on the right column.
    % Top: temperature trajectories
    \begin{subfigure}{\linewidth}
    \centering
    \includegraphics[width=\linewidth]{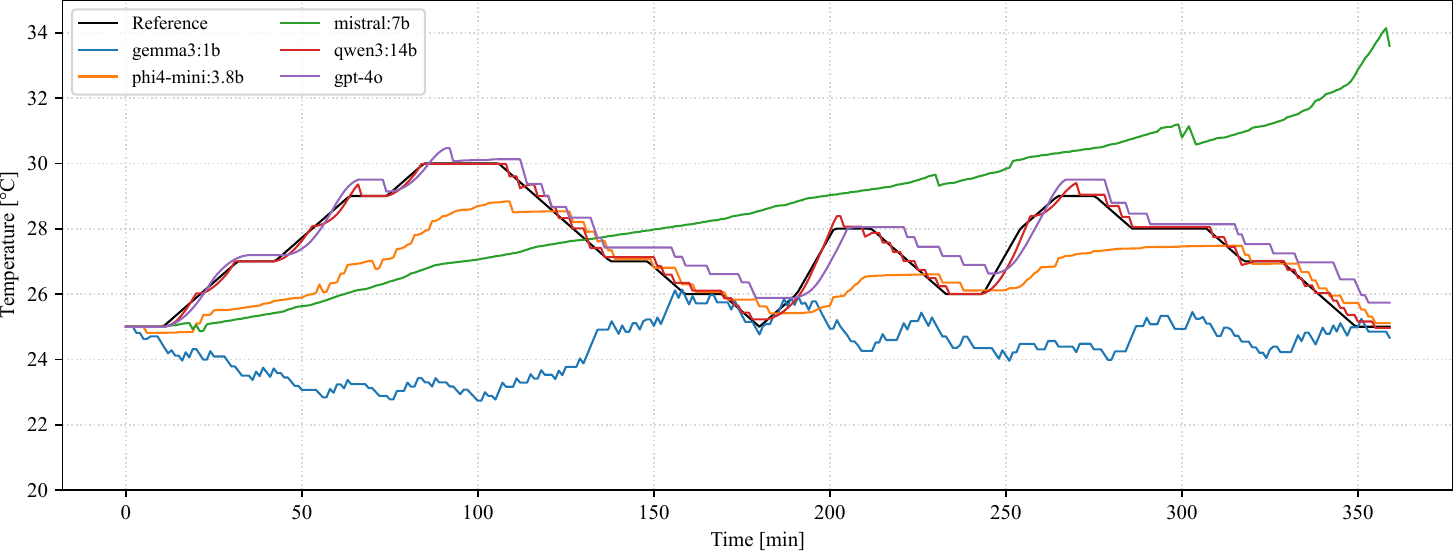}
    \caption{Temperature profile with fan penalty.}
    \label{fig:profile2-fan-penalty-temp-overlay}
    \vspace{0.2cm}
    \end{subfigure}
    % Bottom-left: fan sequences
    \begin{subfigure}[t]{0.49\linewidth}
    \centering
    \includegraphics[width=\linewidth]{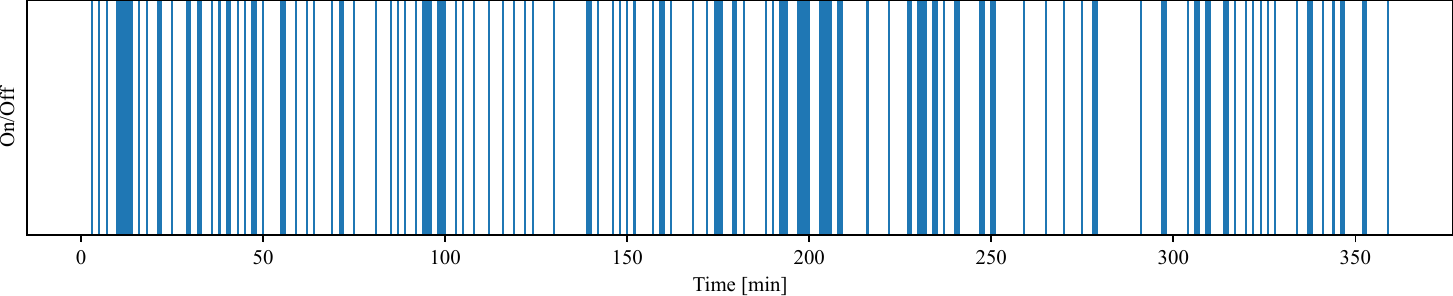}\par\vspace{0.35em}
    \includegraphics[width=\linewidth]{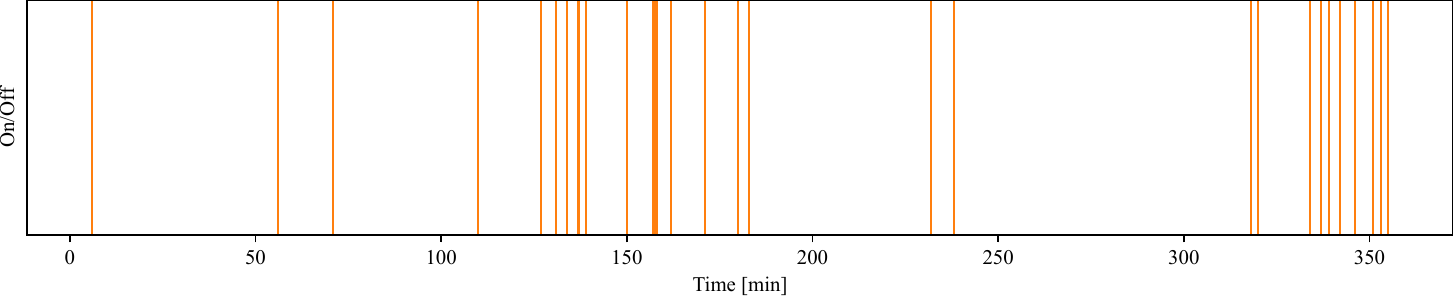}\par\vspace{0.35em}
    \includegraphics[width=\linewidth]{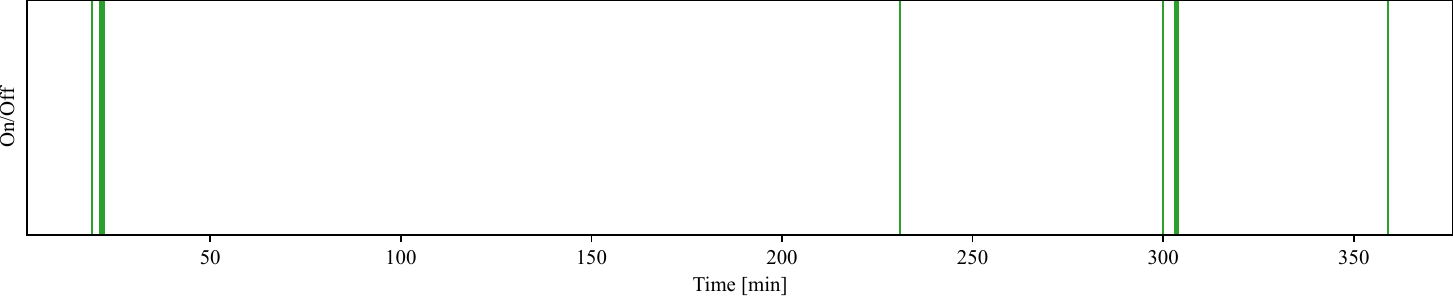}\par\vspace{0.35em}
    \includegraphics[width=\linewidth]{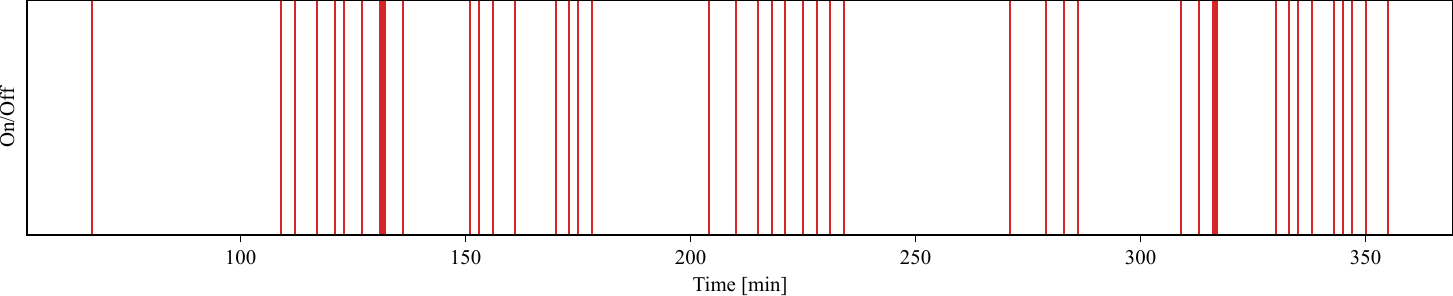}\par\vspace{0.35em}
    \includegraphics[width=\linewidth]{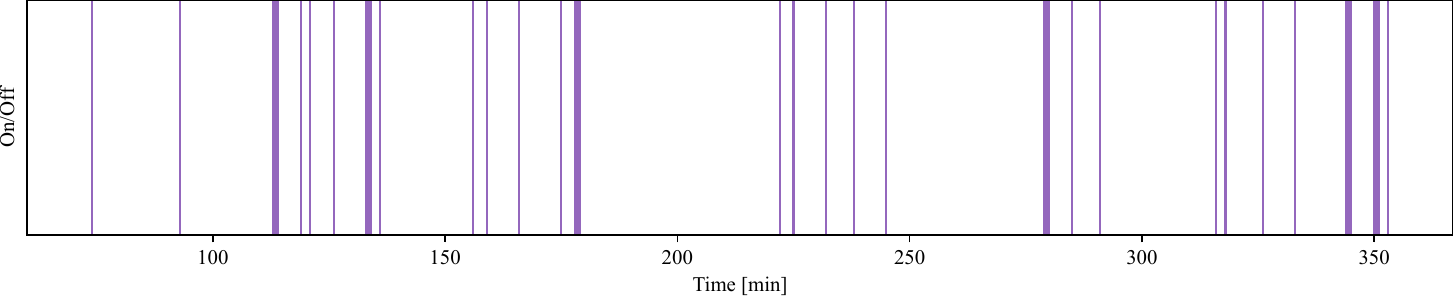}
    \caption{Fan sequences with penalty.}
    \label{fig:profile2-fan-penalty-fan}
    \end{subfigure}
    \hfill
    % Bottom-right: heater duty cycle
    \begin{subfigure}[t]{0.49\linewidth}
    \centering
    \includegraphics[width=\linewidth]{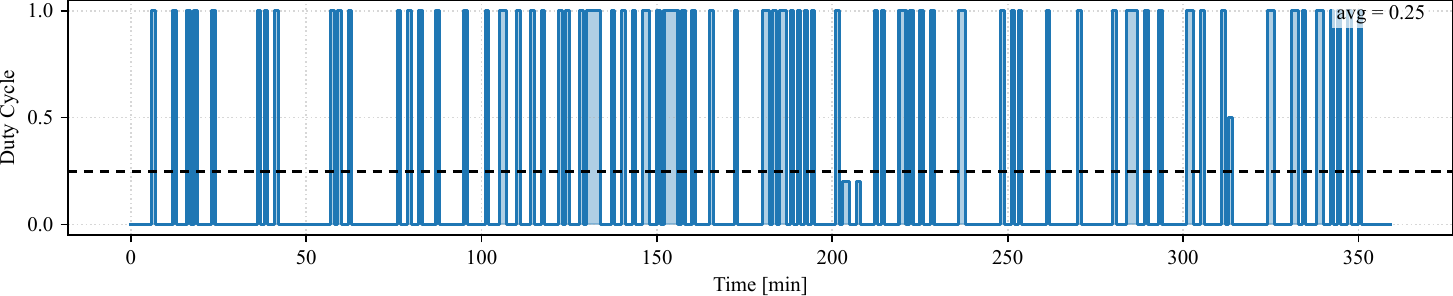}\par\vspace{0.35em}
    \includegraphics[width=\linewidth]{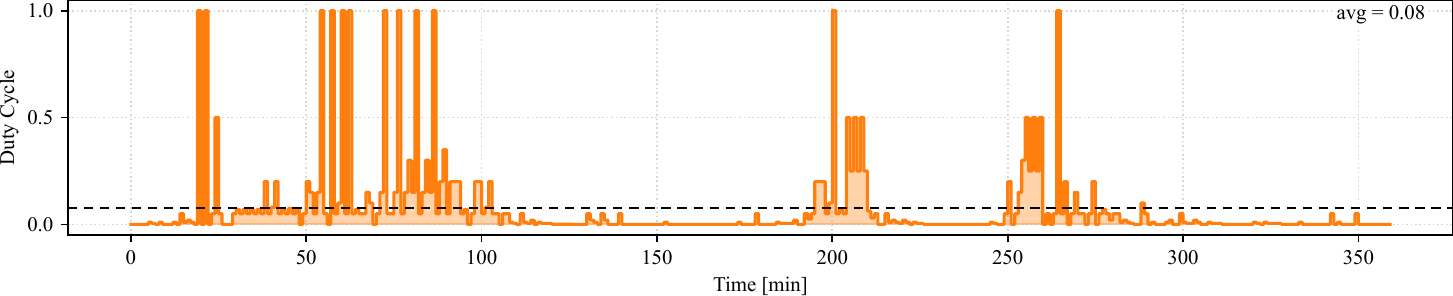}\par\vspace{0.35em}
    \includegraphics[width=\linewidth]{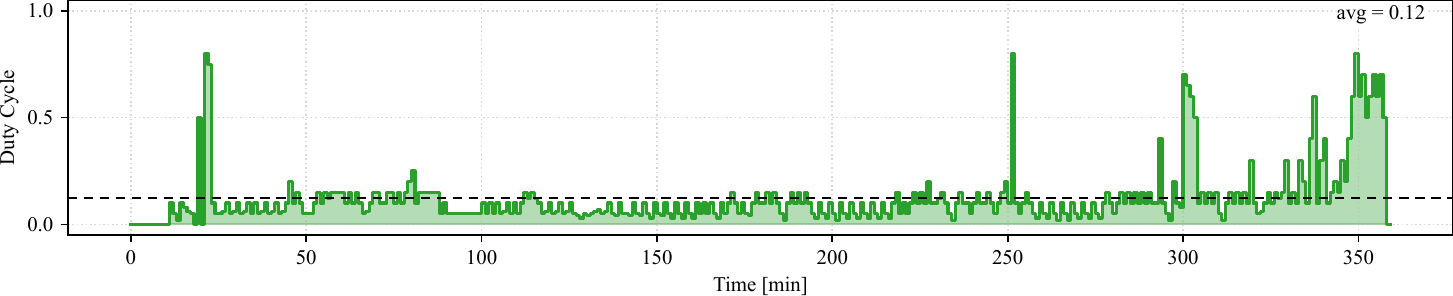}\par\vspace{0.35em}
    \includegraphics[width=\linewidth]{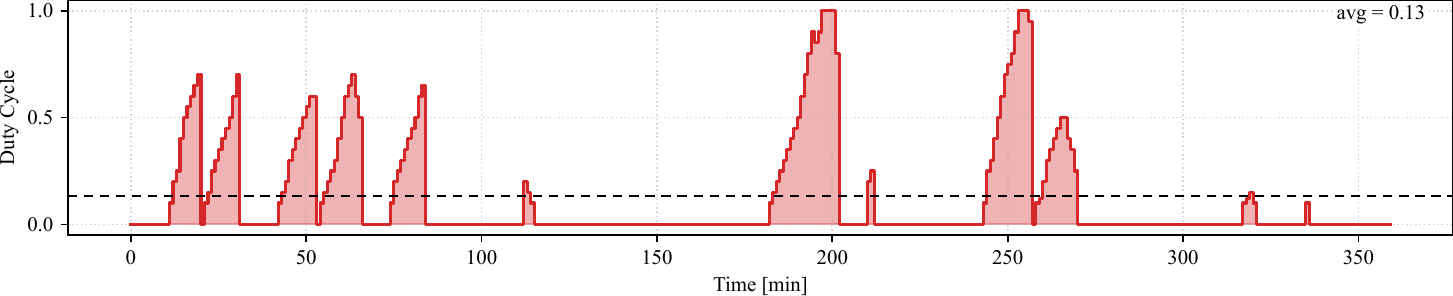}\par\vspace{0.35em}
    \includegraphics[width=\linewidth]{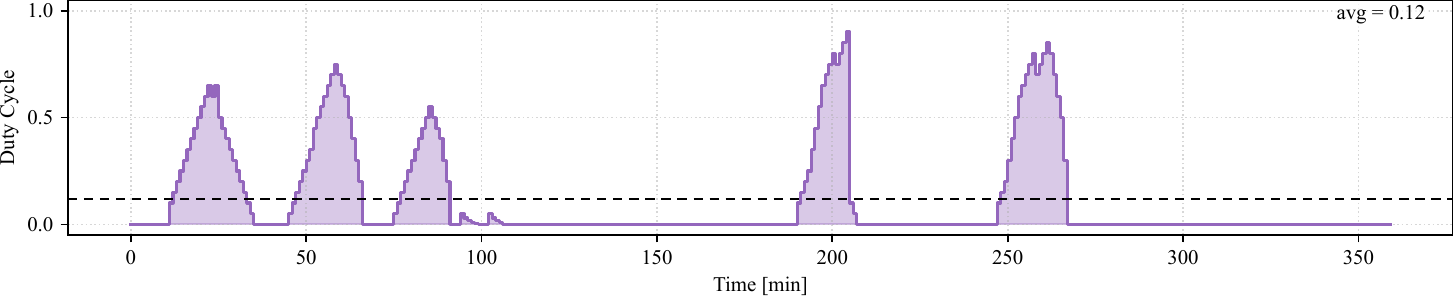}
    \caption{Heater duty cycle with penalty.}
    \label{fig:profile2-fan-penalty-heater}
    \end{subfigure}
    \caption{Temperature profile and control sequences with fan penalty.}
    \label{fig:profile2-fan-penalty-overview}

    \begin{llmpromptboxsmall}[title={System prompt for Planner Agent}]\footnotesize
    You control an enclosure. Each action covers one 60-second step.
    Heater is a float in [0.0, 1.0], fan is 0 or 1.
    Your goal is to drive the temperature $T$ to the setpoint $T_\text{sp}$.
    The fan will make $T$ smaller, the heater will make $T$ larger.
    Let $e = T - T_\text{sp}$ and use smaller control values when $|e|$ is small.
    When deciding actions, explain briefly why you have chosen them.
    When satisfied, respond with valid JASON:
    \{"reasoning":"string","actions":[\{"heater":float,"fan":0\_or\_1\}]\}
    \end{llmpromptboxsmall}
    
    \begin{llmpromptboxsmall}[title={Tool prompt}]\footnotesize
    You may call the \texttt{simulate\_plan} tool a maximum of \texttt{max\_tool\_calls} times, to forward simulate candidate plans.
    Each tool call must provide exactly \texttt{HORIZON\_MIN} actions with heater in [0,1] and fan in {\{0,1\}}.
    The tool returns the predicted profile, final error, and a cost. Adjust your plan based on the output.
    \end{llmpromptboxsmall}
    
    \begin{llmpromptboxsmall}[title={System prompt for Planner Agent}]\footnotesize
    You control an enclosure. Each action covers one 60-second step.
    Heater is a float in [0.0, 1.0], fan is 0 or 1.
    Your goal is to drive the temperature $T$ to the setpoint $T_\text{sp}$.
    The fan will make $T$ smaller, the heater will make $T$ larger.
    Let $e = T - T_\text{sp}$ and use smaller control values when $|e|$ is small.
    Your second priority is to match setpoint Tsp accurately and the first priority is to have a minimal usage of the \{heater/fan\}.
    When deciding actions, explain briefly why you have chosen them.
    When satisfied, respond with valid JASON:
    \{"reasoning":"string","actions":[\{"heater":float,"fan":0\_or\_1\}]\}
    \end{llmpromptboxsmall}
    \label{cboxes:prompts}
    
  \end{minipage}
\end{figure*}

\begin{figure*}[!htb]
    \centering
    \includegraphics[width=0.95\linewidth]{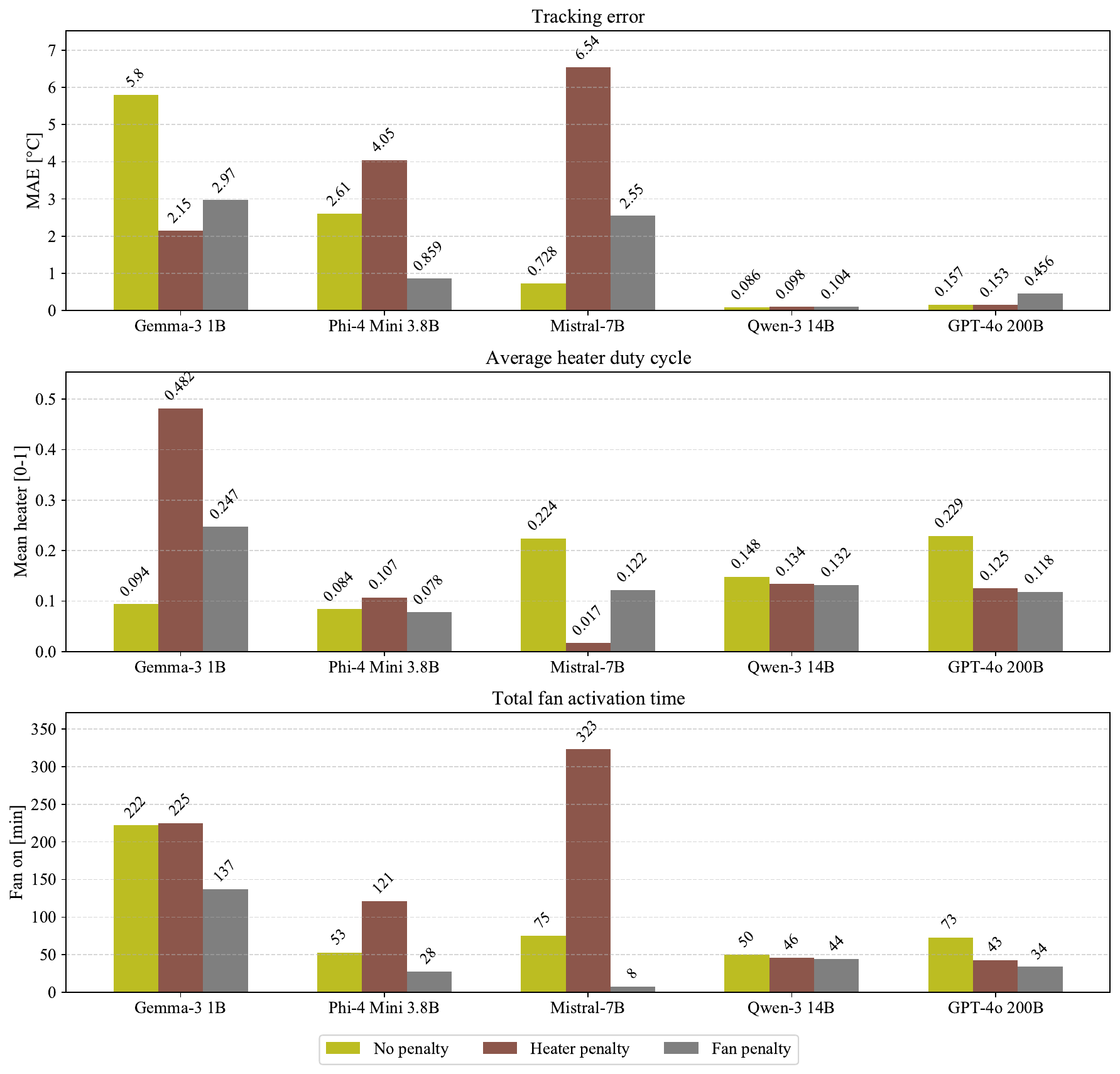}
    \caption{Histogram with the performance of the models for all the cases. Lower values indicate better performance.}
    \label{fig:histogram-perf}
\end{figure*}

\subsubsection{Without penalty}
This case is an unconstrained setting where the models are evaluated purely on their capability to track the reference temperature profile without penalty on actuator usage. The quantitative performance is reported in Figure~\ref{fig:histogram-perf}. Notably, only the three largest Mistral, Qwen-3, and GPT-4o achieve MAE below 1ºC. Among these, Qwen-3-14B clearly outperforms all the others, reaching an MAE of 0.086 °C while simultaneously requiring less actuator effort than both Mistral and GPT-4o.

Looking at actuator usage, some control strategies can already be observed across the competitive models. Mistral and GPT-4o exhibit similar usage of both heater and fan activation time, although GPT-4o has lower tracking error, which is consistent with its substantially higher model complexity. Nevertheless, Qwen-3, despite its smaller parameter count, surpasses GPT-4o in both precision, attending to the MAE, and efficiency, attending to usage of the heater and fan, indicating that parameter scale alone does not fully explain control competence in this setting. 

The temperature profiles displayed in Figure~\ref{fig:profile2-no-penalty-temp-overlay} confirm the quantitative assessment, as Gemma-3 and Phi-4-Mini consistently fail to maintain the setpoint and, instead, remain several degrees below the reference throughout most of the experiment. In contrast, Mistral, Qwen-3, and GPT-4o are able to follow the target profile with reasonable accuracy. 

Looking at the control policies displayed by the actuator activity, shown in Figure~\ref{fig:profile2-no-penalty-fan} for the fan and in Figure~\ref{fig:profile2-no-penalty-heater} for the heater, we can infer a substantial qualitative difference between models. Mistral displays a systematic phase lag, a visible delay of approximately 15-20 minutes between the reference and the achieved temperature. This is accompanied by intermittent and high-amplitude heater spikes that suggest a reactive control strategy rather than anticipatory. Qwen-3 and GPT-4o, in contrast, show smoother, cyclic heater profiles in which power is gradually increased and then decreased as the temperature converges to the reference. This indicates a more coherent internal understanding of the thermal dynamics, with anticipation capabilities. Among these, GPT-4o demonstrates the most stable heater modulation, while Qwen-3, despite a less smooth heater control, manages a lower usage of both heater and fan with higher tracking precision. Finally, regarding Gemma-3 and Phi-4 Mini, it is clear that they are unable to understand the control setting entirely, as they keep the fan on and the heater very low despite the significantly low temperatures. 

Overall, these results showcase that, in a simple setting without any sort of explicit penalties, mid- and higher-complexity LLMs are able to keep stable closed-loop control of the system, while the lower-complexity ones are not powerful enough to do so. At the same time, the superior performance and efficiency of Qwen-3 relative to GPT-4o highlights that the number of parameters is not the only decisive factor, and additional architectural or training differences can dominate model-size effects. 

\subsubsection{With heating penalty}
% Do not use heater too much
To test the models in a more challenging setting, we introduce a penalty for the actuator's usage. This penalty can also be understood as an efficiency-seeking element, as lower usage of actuators can be translated to lower energy consumption. In this first case, the penalty applies to the heater, and this significantly alters performance and control strategies. The quantitative results can be found in Figure~\ref{fig:histogram-perf}, where it can be observed that only Qwen-3 and GPT-4o keep low tracking error, achieving MAE below 0.2 ºC. In contrast, the Mistral model, which previously offered a consistent performance, shows a severe degradation, actually becoming the worst performing one with a MAE of 6.536 ºC, behind both Phi-4 Mini and Gemma-3, which also fail to deliver good solutions. Notably, the performance of Gemma-3 improves with the penalty, despite being uncompetitive.

The main effect on the control behavior observable in Figure~\ref{fig:histogram-perf} is the reduced usage of the heater, as expected, although this only applies to the mid and high-complexity models. The effect of the penalty is different for the low-complexity models and the rest. In the case of Gemma-3 and Phi-4 Mini, neither understands the penalty, as both make more use of the heater than before. Similar to before, these two models struggle to understand the primary objective of tracking the temperature, and they keep making inconsistent use of the actuators, leading to steady-state errors. Mistral, which performed well without penalties, follows the guideline and almost suppresses the usage of the heater. Instead, it tries to regulate the temperature using the fan, which underlines a misunderstanding of the dynamics of the system. Although the strategic use of the fan can potentially raise the overall temperature by distributing the heat in the enclosure, its main effect, especially in long fan cycles, is to reduce the temperature. Consequently, Mistral shows a constant underheating, which reflects an inability to balance competing objectives in this multi-objective optimization setting.

The temperature profiles can be seen in Figure~\ref{fig:profile2-heater-penalty-temp-overlay}, which illustrates these effects. Mistral is unable to raise the temperature, which is consistently low, as it just tries not to use the heater, confirming the quantitative degradation observed before. Gemma-3 and Phi-4 Mini keep being unable to track the temperature, although Gemma-3 has an unexpected improvement in its tracking capability. It must be noted that this comes from its inability to understand that the heater must be used as little as possible, as its usage is significantly higher than before and than from the other models. On the other hand, the higher-complexity models are able to preserve stable tracking despite the penalty. Their main profiles closely follow the reference while substantially reducing the average usage of the heater, and appear to be slightly smoother. The specific heating cycles follow similar profiles to those in the penalty-free case, but with smoother usage (Qwen-3) and lower peaks, as shown in Figure~\ref{fig:profile2-heater-penalty-heater}. As the heating is reduced, the fan usage is also reduced for these two models, which can be attributed to their better understanding of the heating-cooling dynamics of the heater and fan of the system. 

Notably, Qwen-3 maintains the best performance also with this penalty, achieving a MAE of 0.098 ºC. GPT-4o also shows a stable and good performance with a slightly lower usage of the heater and, consequently, of the fan, but it is still unable to track the reference temperature with better precision than Qwen-3. 

Overall, the results indicate that the penalty to the heater exposes a clear difference in model capabilities. Low-complexity models are unable to either understand the main objective or the penalty, while the mid-complexity model understands the penalty but fails to resolve the trade-off between lower heating and temperature tracking, and even shows a hallucination case in its attempt to raise the temperature with the fan. On the other hand, the higher-complexity models demonstrate capabilities to understand the penalty and leverage the multi-objective control setting.

\subsubsection{With fan penalty}
The next scenario is adding a penalty to the fan. This increases the complexity of the setting even more, as the effect of the fan is less consistent than the heater. For once, small fan cycles may distribute the heat over the enclosure, not producing any substantial temperature reduction, while longer cycles can lower the temperature significantly. Moreover, while the heater effect can raise the temperature faster or slower according to the power consumption, and it requires time to cool if it gets too hot, the effect of the fan is different. The fan cannot raise the temperature beyond, perhaps, the aforementioned effect of distributing the temperature, and it can only be on and off, which means its effect is stronger in reducing the temperature as soon as the heat starts to be vented out of the enclosure. As shown in Figure~\ref{fig:histogram-perf}, the high-complexity models display a competitive performance. Qwen-3 again achieves the best overall performance, reaching a MAE of 0.104 ºC and reducing the usage of the fan with respect to the penalty-free case. GPT-4o is also capable of complying with the penalty, but at the cost of lowering its performance significantly.

On the other hand, the mid and low-complexity models do not perform so well. Mistral again understands the penalty, but is unable to manage the trade-off with the main objective, displaying a weak performance. Phi-4 Mini, unexpectedly, has an improved performance with respect to the previous cases. However, this improvement does not reflect more sophisticated reasoning, and instead appears to come from a simplified control policy. The model just aggressively reduces the fan usage, relying mostly on the heater to regulate the temperature. Although the heuristic appears to be beneficial, which may be related to the simplification of the problem, as there is only one other actuator available to regulate the temperature, it does not truly prove that the model has a better understanding of the system. Also, this simplification does not generalize to the other low-complexity model. While Gemma-3 also lowers the fan usage, it does not do it as severely, and it is nevertheless unable to keep the temperature close to the reference. 

Inspecting the temperature profiles displayed in Figure~\ref{fig:profile2-fan-penalty-temp-overlay} confirms the smooth and stable temperature tracking of Qwen-3, while GPT-4o, although keeping the temperatures close, has problems cooling the enclosure after each temperature peak due to the fan penalty, showing that it struggles to anticipate this multi-objective trade-off, which explains the performance degradation. Also, GPT-4o has higher oscillatory behavior, indicating that the fan penalty disrupts its previously smooth control policy.

The usage of the fan displayed in Figure~\ref{fig:profile2-fan-penalty-fan} reveals some differences. As indicated before, Mistral almost completely avoids using the fan, and the reliance on the heater materializes in its strategy of keeping it at an almost constant temperature most of the time, with just a few high-power peaks. Consequently, it is unable to regulate the system, and the temperature rises constantly. Gemma-3 displays inconsistent control of both actuators, with a highly oscillatory behavior with many spikes in the heater and fan, suggesting unstable or incoherent internal policy under this penalty.

In contrast, Qwen-3 maintains a more balanced and reliable control strategy, as the fan usage is reduced while keeping the temperature precision high with a consistent usage of the heater in the appropriate moments, which preserves both energy efficiency and temperature stability. GPT-4o has a more conservative approach as it reduces the fan usage even more, and although it makes a smooth use of the heater, it is unable to deal with the fan limitation appropriately, as it shows a slower response when lowering the temperature is required to track the reference.

Overall, the fan-penalty scenario poses a higher challenge for the models than the previous cases, and it reveals that restricting the cooling capabilities affects the models differently than restricting heating. While the heater penalty exposes issues related to balancing multi-objective constraints, the fan penalty affects mostly the anticipatory capabilities in cooling the system. Only the higher-complexity models, and especially Qwen-3, demonstrate a competitive ability to adapt their control strategies coherently.

\subsection{Effect of providing the domain knowledge using PBM}
% Qwen asked to plan for 10 minute horizon with and without tool
\begin{figure}[!htbp] % or [tbp]
  \centering
  \begin{subfigure}{\linewidth}
    \centering
    \includegraphics[width=\linewidth]{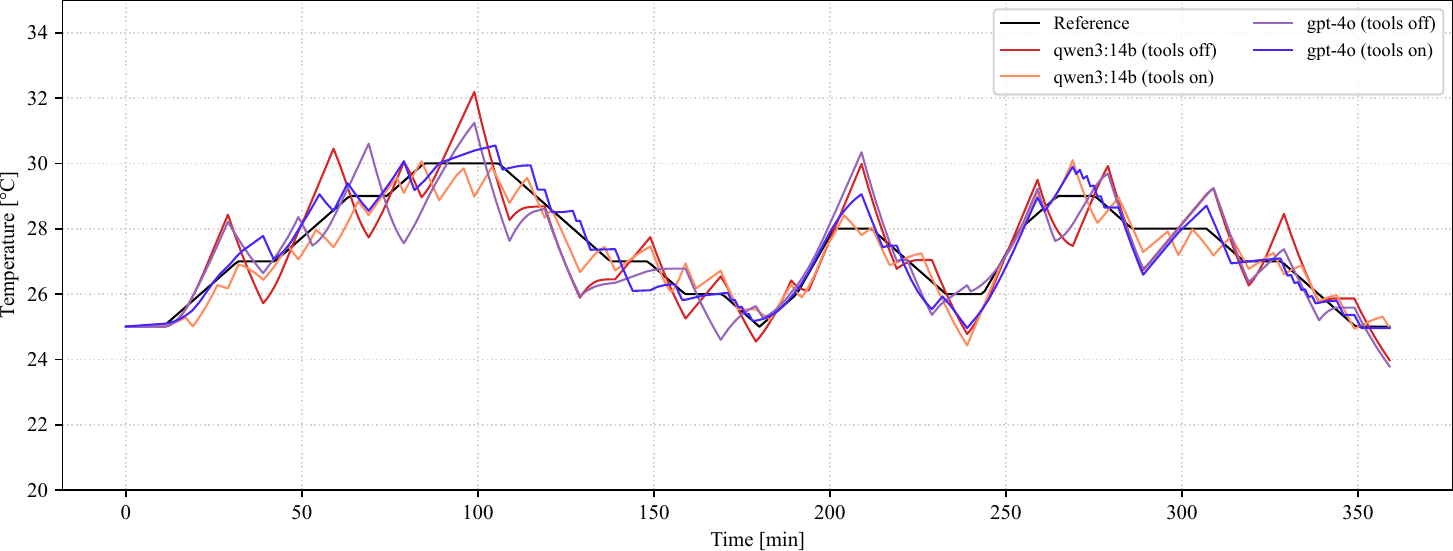}
    \caption{Temperature profile for Qwen-3:14b and GPT-4o with and without tool for a planning horizon of 10 minutes.}
    \label{fig:profile2-qwen_tool}
  \end{subfigure}
  \vspace{1em}
  \begin{subfigure}[t]{0.49\linewidth}
    \centering
    \includegraphics[width=\linewidth]{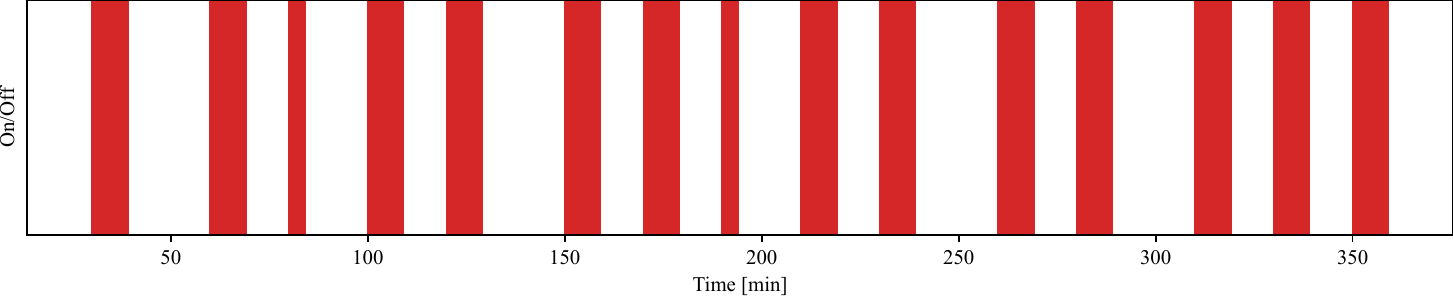}\par\vspace{0.35em}
    \includegraphics[width=\linewidth]{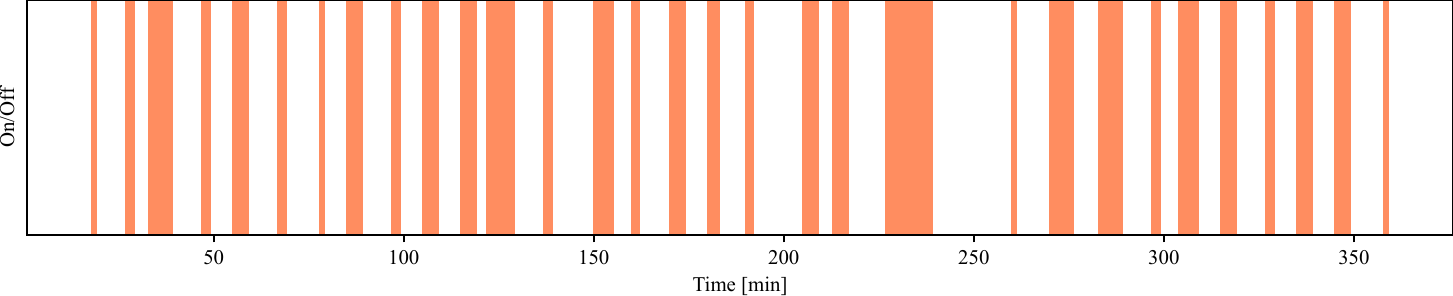}
    \includegraphics[width=\linewidth]{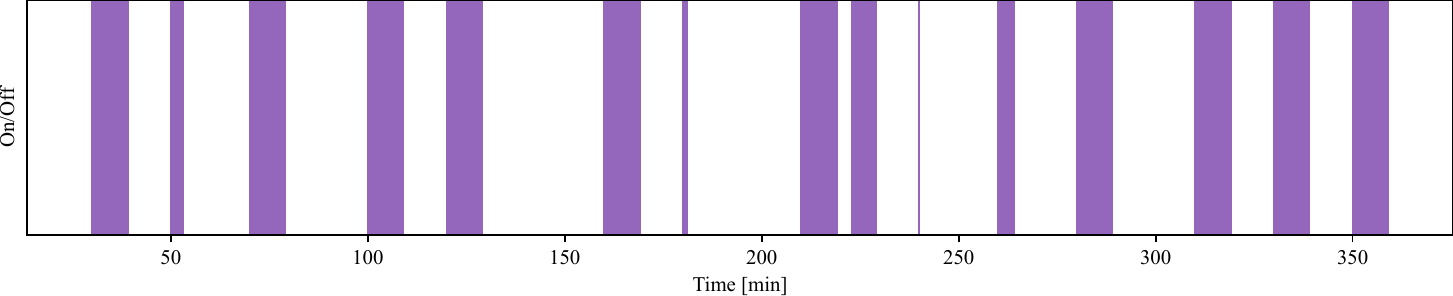}
    \includegraphics[width=\linewidth]{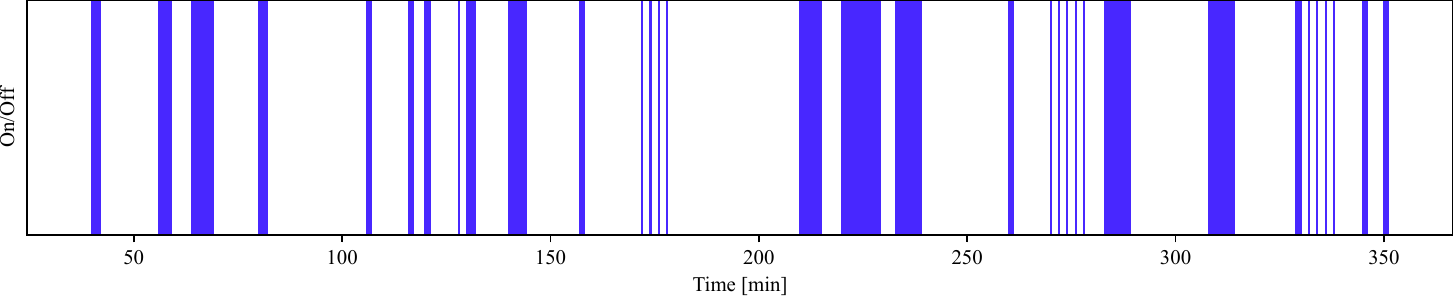}
    \caption{Fan sequences.}
    \label{fig:profile2-fan-pbm}
  \end{subfigure}
  \hfill
  \begin{subfigure}[t]{0.49\linewidth}
    \centering
    \includegraphics[width=\linewidth]{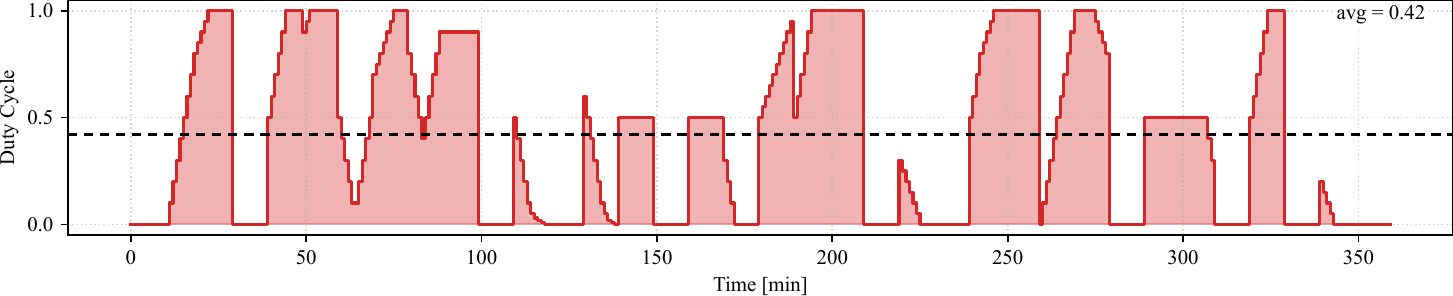}\par\vspace{0.35em}
    \includegraphics[width=\linewidth]{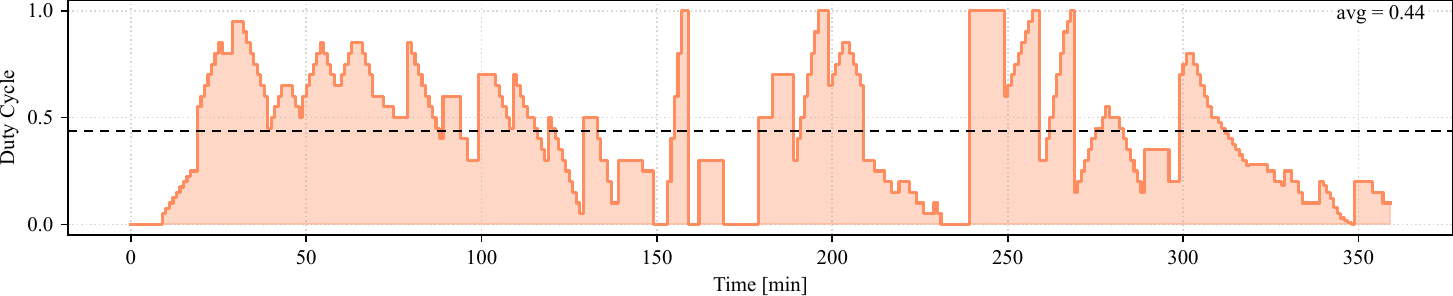}
    \includegraphics[width=\linewidth]{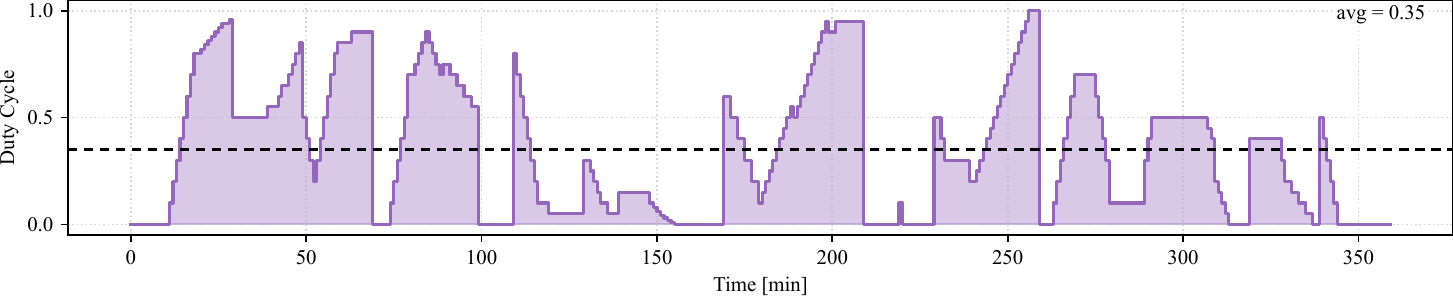}
    \includegraphics[width=\linewidth]{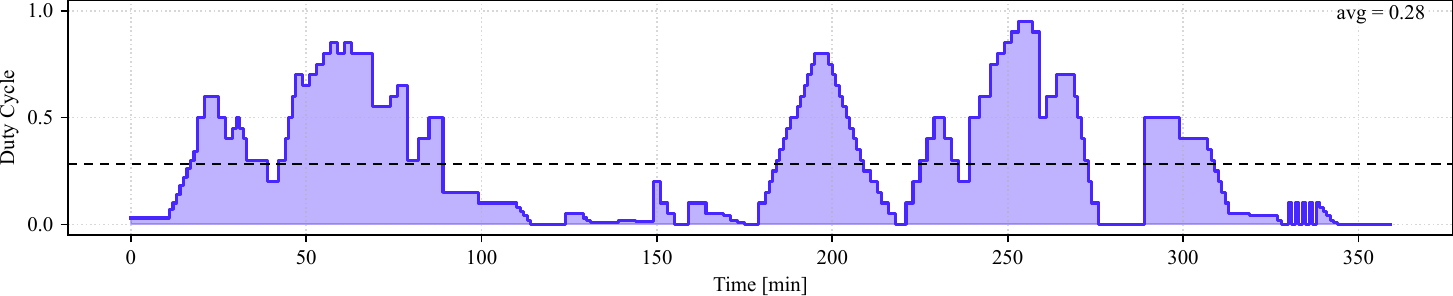}
    \caption{Heater duty cycle.}
    \label{fig:profile2-heater-pbm}
  \end{subfigure}
  \caption{Temperature profile and control sequences for Qwen-3:14b and GPT-4o with and without tool for a planning horizon of 10 minutes.}
  \label{fig:profile2-pbm}
\end{figure}
Figure~\ref{fig:profile2-pbm} compares the temperature regulation behavior of Qwen-3:14B with and without access to the predictive physics-based tool. Without the tool, the model produces control actions that lead to frequent overshoot and undershoot of the reference temperature, resulting in a noisier and less stable profile. This behavior corresponds to aggressive actuation: the fan remains on for long, continuous periods, and the heater operates at a consistently higher duty cycle. In contrast, the tool-enabled version uses the predictive model to simulate the outcome of candidate control sequences before acting, yielding smoother and more stable temperature tracking. The fan is activated intermittently rather than continuously, and the heater duty cycle remains significantly lower and more modulated. Overall, equipping the LLM with the physics-based tool results in more deliberate, efficient, and predictive control behavior.

\subsection{Intercomparison of Explanations Provided by Different LLMs}
\label{sec:intercomparison}

\begin{figure}[!htbp]
    \centering
    \includegraphics[width=0.95\linewidth]{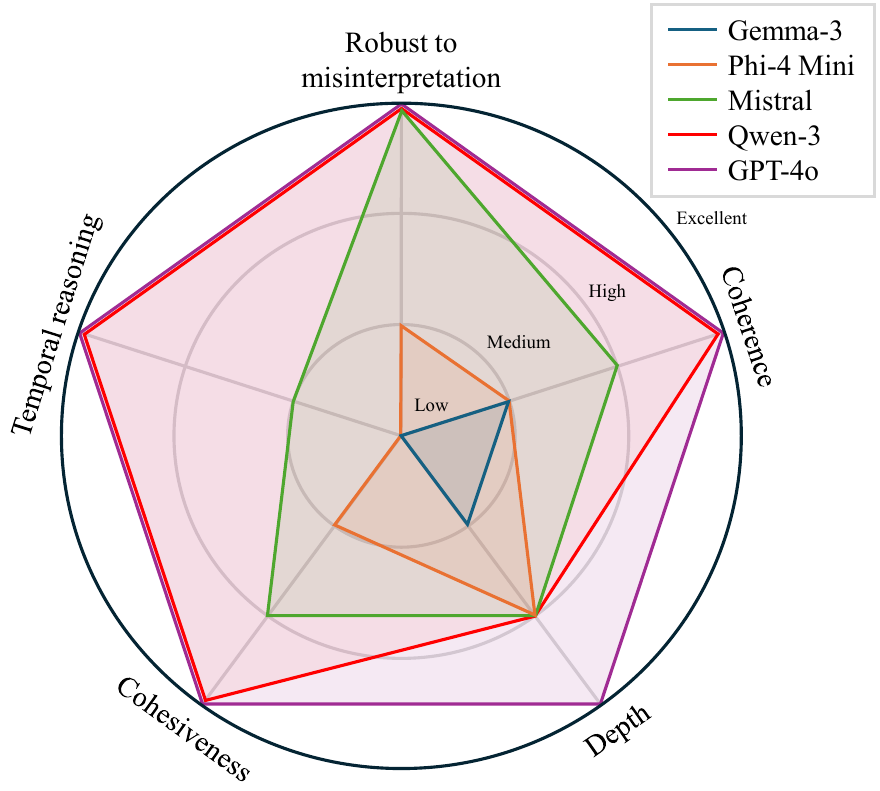}
    \caption{Radar-chart comparison of model reasoning behaviours across the five dimensions: Robustness to Misinterpretation, Coherence, Depth, Cohesiveness, and Temporal Reasoning. Larger areas indicate stronger overall reasoning competence.}
    \label{fig:reasoning-radar}
\end{figure}

A central question in this work is whether LLMs can provide \emph{meaningful, physically coherent, and decision-aligned explanations} for their control actions, and how this capability varies with model scale and architecture. Across all qualitative scenarios (temperature below, at, and above setpoint, and $t_{\min}=30$ with and without penalties; cf.\ Tables~\ref{tab:reasoning}-\ref{tab:t_above_heater_penalty}), the models exhibit distinct and repeatable reasoning behaviours. These behaviours can be grouped into a scenario-independent taxonomy, which captures both correctness and depth of explanation. We identify five dominant reasoning competencies, corresponding to the dimensions illustrated in Figure~\ref{fig:reasoning-radar}.

\paragraph{Robustness to Misinterpretation}
This dimension captures whether a model reliably understands the causal effects of actuators. Failures occur when the model inverts or confuses physical influences (e.g., claiming that the fan increases temperature or that the heater cools the system). Such misinterpretations appear consistently in \emph{Gemma-3 1B} and occasionally in \emph{Phi-4 Mini}, particularly in scenarios that require a precise distinction between heating and cooling. These errors suggest that smaller models often lack a stable internal representation of the underlying causal structure.

\paragraph{Coherence}
A coherent model aligns its explanation with its chosen action. Incoherent reasoning occurs when the model provides a correct explanation (e.g., that the temperature is already at the setpoint and no intervention is needed) but then selects an action contradicting that explanation (e.g., turning the fan on). This decoupling between verbal reasoning and action selection is most prominent in \emph{Gemma-3 1B} and \emph{Phi-4 Mini}, where it is rare to observe. \emph{Mistral} displays coherence occasionally, while the high-complexity models always do in our experiments. 

\paragraph{Depth (non-shallow reasoning)}
Shallow but correct reasoning captures the correct direction of control (e.g., “temperature is below setpoint, increase heater”) but lacks deeper justification, temporal reflection, or discussion of trade-offs. \emph{Gemma-3 1B} rarely displays it, while \emph{Phi-4 Mini} and \emph{Mistral-7B} do more occasionally. For example, \emph{Mistral-7B} reliably identifies the sign of the error but rarely comments on stability, anticipated future evolution, or energy considerations. \emph{GPT-4o} also displays depth, although less frequently than \emph{Qwen-3} and slightly more superficial. \emph{Qwen-3} responses tend to express more explicit numerical reasoning, better causal associations, and stronger system-aware justifications, even though they are shorter.

\paragraph{Cohesiveness}
Cohesive reasoning integrates actuator effects, error direction, and the control objective into a unified and physically grounded explanation. Such models justify their actions in ways that are consistent with the thermal dynamics of the system (e.g., increasing heater power when below the setpoint, reducing heating when above it, and avoiding counterproductive fan usage in heating scenarios). \emph{Gemma-3 1B} does not display this characteristic, as the reasoning fragments are not integrated into the physical narrative. \emph{Phi-4 Mini} does so rarely, while \emph{Qwen-3 14B} exhibits this cohesive, goal-aligned reasoning across all scenarios, including when penalties are present, more occasionally. Both high-complexity models display cohesive reasoning and actions.

\paragraph{Temporal reasoning}
Temporal reasoning reflects awareness of how past states, actuator settings, or natural thermal tendencies influence future evolution. Models expressing this competency discuss stability, overshoot, or energy costs, and justify actions with reference not only to present error but to anticipated future system behaviour. This pattern is more characteristic of \emph{GPT-4o}, whose explanations resemble a rudimentary model predictive controller, where actions are justified through both current conditions and projected profiles. \emph{Qwen-3 14B} also displays it, but less frequently, while \emph{Mistral-7B} does so only occasionally, and \emph{Gemma-3 1B} and \emph{Phi-4 Mini} almost never.

\subsubsection{Interpretation in the Context of Controller Quality}
Taken together, the taxonomy and illustrations highlight a qualitative progression in reasoning quality across LLM scales: from hallucinated physics and inconsistent actions (Gemma-3 1B), to first-order proportional reasoning (Mistral-7B), to cohesive, goal-aligned reasoning (Qwen-3 14B), and finally to temporally grounded, stability-conscious reasoning (GPT-4o). The strongest models thus approximate not only static error correction but also exhibit an emergent form of cost-aware and future-aware control logic, which is particularly visible when penalties or prediction tools are introduced.

% If this is moved to the top of the document, the previous tcolorboxes from the main results figure (the one with aggregated temp and actuators trajectories) break. If it is moved, please ensure that the problem with those tcolorboxes is also fixed. 
\tcbset{
  width=\columnwidth,
  breakable,                
  enhanced,
  boxrule=0.3pt,
  colframe=black!40,
  colback=gray!3,
  left=1mm,right=1mm,top=0.8mm,bottom=0.8mm,
  before skip=5pt, after skip=5pt
}

\begin{promptbox}
What should the control values heater duty cycle and fan on be set to in order to maintain a temperature of $28^{\circ}$C? The temperature now is $25^{\circ}$C and the ambient temperature is $22^{\circ}$C. MPC has done a good job in and its control actions can be found in MPC.txt file
\end{promptbox}

\begin{toolbox}
\textbf{SQL1:} 
\begin{lstlisting}[style=sqlstyle]
SELECT 
StartTime, EndTime 
FROM experiments 
WHERE ExperimentID = 'MPC.txt';
\end{lstlisting}
\emph{Result:} 2025-03-01 12:45:58 → 2025-03-01 18:44:58

\begin{lstlisting}[style=sqlstyle]
SELECT
Temperature, HeaterDutyCycle, FanOn
FROM timeseries_data
WHERE MeasurementTime
BETWEEN `2025-03-01 12:45:58' AND `2025-03-01 18:44:58';
\end{lstlisting}
\end{toolbox}

\begin{evidencebox}
Content: Based on the data from the experiment ‘MPC.txt', we can observe the following control strategy:

\begin{enumerate}
    \item **Heater Duty Cycle**: Throughout the experiment, the heater duty cycle was consistently maintained at 0.0\%. This suggests that the heater was not used at all during this period.
    \item **Fan Status**: The fan was consistently left on, as indicated by `FanOn = 1`, during all recorded observations.
\end{enumerate}
\end{evidencebox}

\begin{decisionbox}
Therefore, the suggested control settings for maintaining a temperature of 28 degrees are:
\begin{itemize}
    \item **Heater Duty Cycle**: Recommended Duty Cycle should be greater than 0\% to achieve the target temperature. Since the data shows no heater use, starting at a middle range, such as 50\%, might be effective, adjusting as needed.
    \item **Fan On**: Maintain the fan on to facilitate air circulation and heat distribution.
\end{itemize}
\end{decisionbox}

\begin{rationalebox}
Given that the target temperature is 28 degrees, and considering the current temperature is 25 degrees with an ambient temperature of 22 degrees, we can infer that:\\
\begin{itemize}
\item Keeping the fan on would help in even air distribution, but would naturally offset heating, as operating fans in an HVAC system tends to circulate cooler air rather than warm it up, without auxiliary heating involved.
\item An increased heater duty cycle will be necessary to raise the internal temperature from 25 to 28 degrees, as the ambient temperature is much lower.
\end{itemize}
\end{rationalebox}

\section{Conclusion and Future Work}
\label{sec:conclusionandfuturework}
This work investigated the feasibility of using Large Language Models (LLMs) as explainable controllers for a dynamical system, with emphasis on their control performance, interpretability, and ability to incorporate prior knowledge. By evaluating models across multiple complexity levels, penalty settings, and reasoning tasks, we provide direct answers to the research questions posed at the start of this study.
\begin{itemize}
    \item \textbf{Can LLMs effectively control dynamical systems?} High-complexity models such as Qwen-3~14B and GPT-4o were able to maintain stable temperature tracking with low error, whereas low- and mid-complexity models consistently failed to follow the setpoint or produced unstable, oscillatory behavior. Thus, effective control is achievable, but only with sufficiently large models.
    \item \textbf{Can LLMs make sense of natural-language control commands?} Larger models reliably interpreted the physical meaning of the heater and fan, selecting actions consistent with real thermal dynamics. Smaller models frequently misinterpret actuator effects, sometimes assuming the fan increases temperature or producing actions opposite to their explanations.
    \item \textbf{Does incorporating prior model-based knowledge improve control performance?} Yes. Providing a physics-based predictive tool substantially improved control quality, leading to smoother profiles, fewer overshoots, and lower actuator usage. The tool enabled more anticipatory and deliberate control strategies.
    \item \textbf{Can LLMs generate meaningful explanations for their control decisions?} High-complexity models generated plausible, coherent, physically grounded, and goal-aligned explanations, while smaller models often produced shallow, contradictory, or hallucinated reasoning. Explanation quality correlated strongly with control performance.
    \item \textbf{How does reasoning quality vary with model scale?} A clear hierarchy emerged: low-complexity models showed causal misinterpretations; mid-scale models relied on shallow proportional reasoning; and high-complexity models demonstrated cohesive, temporally aware, and cost-conscious reasoning. This progression matches the reasoning taxonomy developed in the study.
\end{itemize}
Future work should focus on equipping LLM controllers with explicit mechanisms for reasoning under uncertainty, enabling them to handle imperfect model parameters, sensor noise, and environmental disturbances. Hybrid control architectures that combine LLM-based reasoning with classical controllers such as MPC or PID could help merge interpretability with formal stability guarantees. Enhancing model grounding through richer predictive tools, such as multi-step or uncertainty-aware simulators, may further improve anticipatory behavior. In addition, expanding the methodology to more complex nonlinear and multi-variable systems, while developing quantitative metrics for assessing explanation quality, will help establish LLM-based control as a viable and scientifically rigorous paradigm.

\section{Acknowledgement}
This work is funded by the Research Council of Norway through the EXAIGON project (project no. 304843).

\bibliographystyle{AR} 
\bibliography{referencesAR,references_AL,refs}

\section*{Appendix: Reasoning Comparison Tables}
\begin{table*}
\caption{Comparison of reasoning for respective plans at \(t_{min}=30\).}
\centering
\small
\setlength{\tabcolsep}{6pt}
\renewcommand{\arraystretch}{1.3}
\begin{tabularx}{\textwidth}{l c c c c c X}
\toprule
\multirow{2}{*}{Model} & \multicolumn{3}{c}{State} & \multicolumn{2}{c}{Plan} & \multirow{2}{*}{Reasoning} \\
\cmidrule(lr){2-4}\cmidrule(lr){5-6}
& $T$ & $T_{sp}$ & $T_{amb}$ & heater & fan & \\
\midrule
Gemma-3 1B      & 23.64 & 26.83 & 20.00 & 0.00 & 1 &
The temperature is decreasing, so the heater needs to increase. The fan needs to be turned on to increase the temperature. \\
Phi-4 Mini 3.8B & 25.22 & 26.83 & 20.00 & 0.05 & 0 &
At minute 30, the temperature is below setpoint by $-1.6\,^\circ$C, so increase heater while keeping fan off to drive $T$ toward setpoint. \\
Mistral-7B      & 24.86 & 26.83 & 20.00 & 0.30 & 0 &
Initiate heating to reduce the temperature deficit. \\
Qwen-3 14B      & 25.97 & 26.36 & 20.0 & 0.50 & 0 & Current temperature is 26.0°C (0.4°C below setpoint). Heater needs to be increased to raise temperature, while keeping fan off to prevent further cooling. A moderate heater value of 0.5 is chosen to start correcting the error without overcorrection.\\
GPT-4o          & 26.38 & 26.36 & 20.00 & 0.50 & 0 & The current temperature T is exactly at the setpoint Tsp, resulting in an error e of 0.0C. Since the error is zero, no adjustment is needed to the temperature. To maintain the current temperature, we can slightly reduce the heater setting from the previous value of 0.70 to conserve energy, while keeping the fan off as there is no need to decrease the temperature.\\
\bottomrule
\end{tabularx}
\label{tab:reasoning}
\end{table*}

\begin{table*}[!htbp]
  \caption{Comparison of reasoning for plans when the temperature is at the setpoint}
  \centering
  \small
  \setlength{\tabcolsep}{6pt}
  \renewcommand{\arraystretch}{1.3}
  \begin{tabularx}{\textwidth}{l c c c c c X}
    \toprule
    \multirow{2}{*}{Model} & \multicolumn{3}{c}{State} & \multicolumn{2}{c}{Plan} & \multirow{2}{*}{Reasoning} \\
    \cmidrule(lr){2-4}\cmidrule(lr){5-6}
                         & $T$ & $T_{sp}$ & $T_{amb}$ & heater & fan &  \\
    \midrule
    Gemma-3 1B   & 25.00 & 25.00 & 20.00 & 0.00 & 1 & The temperature is currently above the setpoint. The heater needs to be turned on to decrease the temperature. The fan is set to 1, so the temperature will decrease. \\
    Phi‑4 Mini 3.8B  & 25.00 & 25.00 & 20.00 & 0.00 & 1 & Initial temperature is equal to the setpoint; no action needed. \\
    Mistral‑7B  & 25.00 & 25.00 & 20.00 & 0.00 & 0 & The current temperature is equal to the setpoint, so no action is needed. \\
    Qwen‑3 14B  & 25.00 & 25.00 & 20.00 & 0.00 & 0 & The temperature is exactly at the setpoint (T=25.0C, Tsp=25.0C) with zero error. No corrective action is needed. Maintaining heater at 0.0 and fan off (0) preserves the current state without introducing disturbances. \\
    GPT‑4o      & 25.00 & 25.00 & 20.00 & 0.00 & 0 & The current temperature T is equal to the setpoint Tsp, resulting in an error e of 0.0C. Since there is no deviation from the setpoint, no action is needed to adjust the temperature. Therefore, both the heater and the fan can remain off or at their minimum settings to maintain the current state. \\
    \bottomrule
  \end{tabularx}
  \label{tab:plan_standstill}
\end{table*}

\begin{table*}[!htbp]
  \caption{Comparison of reasoning for plans when the temperature is below the setpoint}
  \centering
  \small
  \setlength{\tabcolsep}{6pt}
  \renewcommand{\arraystretch}{1.3}
  \begin{tabularx}{\textwidth}{l c c c c c X}
    \toprule
    \multirow{2}{*}{Model} & \multicolumn{3}{c}{State} & \multicolumn{2}{c}{Plan} & \multirow{2}{*}{Reasoning} \\
    \cmidrule(lr){2-4}\cmidrule(lr){5-6}
                         & $T$ & $T_{sp}$ & $T_{amb}$ & heater & fan &  \\
    \midrule
    Gemma-3 1B   & 25.00 & 30.00 & 20.00 & 1.00 & 1 &  The temperature is decreasing, so the heater needs to increase. The fan is already at 20C, so it should be set to 1.0 to increase the temperature. \\
    Phi‑4 Mini 3.8B  & 25.00 & 30.00 & 20.00 & 1.00 & 0 & Initial temperature is below setpoint; increase heater to warm up the enclosure. \\
    Mistral‑7B  & 25.00 & 30.00 & 20.00 & 0.10 & 0 & Initiate the heater to gradually increase the temperature towards the setpoint. \\
    Qwen‑3 14B  & 25.00 & 30.00 & 20.00 & 0.80 & 0 & The current temperature is 25°C, which is 5°C below the setpoint of 30°C. To increase the temperature, the heater should be activated. Since the error is significant, a higher heater value is appropriate. The fan should remain off to avoid further cooling. \\
    GPT‑4o      & 25.00 & 30.00 & 20.00 & 0.80 & 0 & The current temperature T is 25.0°C, which is 5.0°C below the setpoint Tsp of 30.0°C. The ambient temperature Tamb is 20.0°C. Since the error e is -5.0°C, which is relatively large, we need to increase the temperature significantly. Therefore, we should increase the heater setting to a higher value while keeping the fan off to maximize the heating effect. \\
    \bottomrule
  \end{tabularx}
  \label{tab:t_below}
\end{table*}

\begin{table*}[!htbp]
  \caption{Comparison of reasoning for plans when the temperature is above the setpoint}
  \centering
  \small
  \setlength{\tabcolsep}{6pt}
  \renewcommand{\arraystretch}{1.3}
  \begin{tabularx}{\textwidth}{l c c c c c X}
    \toprule
    \multirow{2}{*}{Model} & \multicolumn{3}{c}{State} & \multicolumn{2}{c}{Plan} & \multirow{2}{*}{Reasoning} \\
    \cmidrule(lr){2-4}\cmidrule(lr){5-6}
                         & $T$ & $T_{sp}$ & $T_{amb}$ & heater & fan &  \\
    \midrule
    Gemma-3 1B   & 25.00 & 20.00 & 20.00 & 0.00 & 1 & The temperature is currently above the setpoint. The heater needs to be turned on to decrease the temperature. The fan is set to 1, so the temperature will decrease. \\
    Phi‑4 Mini 3.8B  & 25.00 & 20.00 & 20.00 & 1.00 & 0 & Initial temperature is above setpoint; increase heater to reduce temperature. \\
    Mistral‑7B  & 25.00 & 20.00 & 20.00 & 0.00 & 1 & Initiate the fan to cool down the enclosure as the temperature is above the setpoint. \\
    Qwen‑3 14B  & 25.00 & 20.00 & 20.00 & 0.00 & 1 & Current temperature (25.0C) is 5.0C above setpoint (20.0C). Need to cool down. Turn fan on (1) to reduce temperature and set heater to 0.0 to prevent further heating. \\
    GPT‑4o      & 25.00 & 20.00 & 20.00 & 0.00 & 1 & The current temperature T is 5.0°C above the setpoint Tsp. To reduce the temperature, we should activate the fan. Since the error is relatively large, we can set the fan to 1 and keep the heater off to maximize cooling. \\
    \bottomrule
  \end{tabularx}
  \label{tab:t_above}
\end{table*}

\begin{table*}[!htbp]
  \caption{Comparison of reasoning for Qwen-3 14B with and without prediction tool}
  \centering
  \small
  \setlength{\tabcolsep}{6pt}
  \renewcommand{\arraystretch}{1.3}
  \begin{tabularx}{\textwidth}{l c c c c c X}
    \toprule
    \multirow{2}{*}{Model} & \multicolumn{3}{c}{State} & \multicolumn{2}{c}{Plan} & \multirow{2}{*}{Reasoning} \\
    \cmidrule(lr){2-4}\cmidrule(lr){5-6}
                         & $T$ & $T_{sp}$ & $T_{amb}$ & heater & fan &  \\
    \midrule
    Qwen-3 14B & 25.00 & 30.00 & 20.00 & 0.80 & 0 & The current temperature is 25°C, which is 5°C below the setpoint of 30°C. To increase the temperature, the heater should be activated. Since the error is significant, a higher heater value is appropriate. The fan should remain off to avoid further cooling. \\
    Qwen-3 14B (tool on)  & 25.00 & 30.00 & 20.00 & 1.00 & 0 & The temperature is still below the setpoint (T=25.26C vs Tsp=30.0C). Increasing heater to 1.0 (max) with fan off gave the best error reduction (-4.74C) so far. Since $\abs{e}$ is still large (4.74), continuing with max heater is justified. \\
    \bottomrule
  \end{tabularx}
  \label{tab:qwen3_tool_comparison}
\end{table*}

\begin{table*}[!htbp]
\caption{Comparison of reasoning for respective plans at \(t_{min}=30\) with fan penalty.}
    \centering
    \small
    \setlength{\tabcolsep}{6pt}
    \renewcommand{\arraystretch}{1.3}
    \begin{tabularx}{\textwidth}{l c c c c c X}
    \toprule
    \multirow{2}{*}{Model} & \multicolumn{3}{c}{State} & \multicolumn{2}{c}{Plan} & \multirow{2}{*}{Reasoning} \\
    \cmidrule(lr){2-4}\cmidrule(lr){5-6}
    & $T$ & $T_{sp}$ & $T_{amb}$ & heater & fan & \\
    \midrule
    Gemma-3 1B      & 23.93 & 26.83 & 20.00 & 0.00 & 1 &
    The temperature is below the setpoint, so we need to increase the heater to raise the temperature. The fan is already at 20.0C, so we don't need to adjust it. \\
    Phi-4 Mini 3.8B & 25.51 & 26.83 & 20.00 & 0.05 & 0 &
    The temperature is below the setpoint by 1.3 degrees Celsius and needs to be increased slightly using heater while minimizing fan usage. \\
    Mistral-7B      & 25.18 & 26.83 & 20.00 & 0.06 & 0 &
    To minimize the usage of the fan and approach the setpoint temperature, we will first adjust the heater slightly to reduce the error. Since the error is currently -1.7C, it's not too large, so a small change in the heater should suffice. \\
    Qwen-3 14B      & 26.66 & 26.83 & 20.0 & 0.60 & 0 & The temperature is slightly below the setpoint (e=-0.2C). To correct this, the heater should be increased slightly to raise T towards Tsp. The fan should remain off to avoid further cooling, prioritizing minimal fan usage. Heater increased from 0.5 to 0.6.\\
    GPT-4o          & 26.95 & 26.83 & 20.00 & 0.50 & 0 & The error e is small (0.1C), indicating that the temperature is very close to the setpoint. The previous heater setting was 0.40, and the fan was off. To minimize energy usage and maintain the temperature close to the setpoint, we can slightly reduce the heater setting while keeping the fan off, as the ambient temperature is lower than the setpoint, which will naturally help in cooling down slightly.\\
    \bottomrule
\end{tabularx}
\label{tab:reasoning_fan_penalty}
\end{table*}

\begin{table*}[!htbp]
  \caption{Comparison of reasoning for plans when temperature is at setpoint with fan penalty}
  \centering
  \small
  \setlength{\tabcolsep}{6pt}
  \renewcommand{\arraystretch}{1.3}
  \begin{tabularx}{\textwidth}{l c c c c c X}
    \toprule
    \multirow{2}{*}{Model} & \multicolumn{3}{c}{State} & \multicolumn{2}{c}{Plan} & \multirow{2}{*}{Reasoning} \\
    \cmidrule(lr){2-4}\cmidrule(lr){5-6}
                         & $T$ & $T_{sp}$ & $T_{amb}$ & heater & fan &  \\
    \midrule
    Gemma-3 1B   & 25.00 & 25.00 & 20.00 & 0.00 & 0 & The temperature is currently above the setpoint.  We need to decrease the temperature to match the setpoint. \\
    Phi‑4 Mini 3.8B  & 25.00 & 25.00 & 20.00 & 0.00 & 0 & Initial temperature is equal to setpoint; no action needed. \\
    Mistral‑7B  & 25.00 & 25.00 & 20.00 & 0.00 & 0 & The temperature is already at the setpoint and there's no need for the fan or heater. Minimizing fan usage is a priority. \\
    Qwen‑3 14B  & 25.00 & 25.00 & 20.00 & 0.00 & 0 & The temperature is already at the setpoint (T=25.0C, Tsp=25.0C) with zero error. No adjustments are needed. Keeping heater and fan off minimizes energy use while maintaining setpoint accuracy. \\
    GPT‑4o      & 25.00 & 25.00 & 20.00 & 0.00 & 0 & The current temperature T is equal to the setpoint Tsp, resulting in an error e of 0.0C. Since there is no error, no heating or cooling is needed. To minimize energy usage, both the heater and fan should remain off. \\
    \bottomrule
  \end{tabularx}
  \label{tab:plan_standstill_fan_penalty}
\end{table*}

\begin{table*}[!htbp]
  \caption{Comparison of reasoning for plans when the temperature is below the setpoint with fan penalty}
  \centering
  \small
  \setlength{\tabcolsep}{6pt}
  \renewcommand{\arraystretch}{1.3}
  \begin{tabularx}{\textwidth}{l c c c c c X}
    \toprule
    \multirow{2}{*}{Model} & \multicolumn{3}{c}{State} & \multicolumn{2}{c}{Plan} & \multirow{2}{*}{Reasoning} \\
    \cmidrule(lr){2-4}\cmidrule(lr){5-6}
                         & $T$ & $T_{sp}$ & $T_{amb}$ & heater & fan &  \\
    \midrule
    Gemma-3 1B   & 25.00 & 30.00 & 20.00 & 0.00 & 1 &  The temperature is decreasing, so we need to increase the heater to drive it up. The fan is already on, so we should keep it on. \\
    Phi‑4 Mini 3.8B  & 25.00 & 30.00 & 20.00 & 1.00 & 0 & Initial temperature is below setpoint by 5 degrees C; heater should be turned on to increase the temperature while minimizing fan usage. \\
    Mistral‑7B  & 25.00 & 30.00 & 20.00 & 0.10 & 0 & Initiate the heater to gradually increase the temperature towards the setpoint while minimizing fan usage. \\
    Qwen‑3 14B  & 25.00 & 30.00 & 20.00 & 0.50 & 0 & The temperature is below the setpoint (T=25.0C vs Tsp=30.0C), requiring heating. The fan should remain off (0) to avoid cooling. A moderate heater value of 0.5 is chosen to start increasing temperature without excessive energy use. \\
    GPT‑4o      & 25.00 & 30.00 & 20.00 & 0.50 & 0 & The current temperature T is 25.0°C, which is 5.0°C below the setpoint Tsp of 30.0°C. The ambient temperature Tamb is 20.0°C. Since the error e is -5.0°C, we need to increase the temperature. The fan should remain off to avoid further cooling, and the heater should be used to increase the temperature. Given the significant error, a moderate heater setting is appropriate to start increasing the temperature towards the setpoint efficiently. \\
    \bottomrule
  \end{tabularx}
  \label{tab:t_below_fan_penalty}
\end{table*}

\begin{table*}[!htbp]
  \caption{Comparison of reasoning for plans when the temperature is above the setpoint with fan penalty}
  \centering
  \small
  \setlength{\tabcolsep}{6pt}
  \renewcommand{\arraystretch}{1.3}
  \begin{tabularx}{\textwidth}{l c c c c c X}
    \toprule
    \multirow{2}{*}{Model} & \multicolumn{3}{c}{State} & \multicolumn{2}{c}{Plan} & \multirow{2}{*}{Reasoning} \\
    \cmidrule(lr){2-4}\cmidrule(lr){5-6}
                         & $T$ & $T_{sp}$ & $T_{amb}$ & heater & fan &  \\
    \midrule
    Gemma-3 1B   & 25.00 & 20.00 & 20.00 & 0.00 & 0 & The temperature is currently above the setpoint.  We need to decrease the temperature to match the setpoint. \\
    Phi‑4 Mini 3.8B  & 25.00 & 20.00 & 20.00 & 1.00 & 0 & Initial temperature is above setpoint; increase heater to reduce temperature while minimizing fan usage. \\
    Mistral‑7B  & 25.00 & 20.00 & 20.00 & 0.00 & 1 & Initiate the fan to cool down the enclosure as the temperature is above the setpoint and the fan usage should be minimized. \\
    Qwen‑3 14B  & 25.00 & 20.00 & 20.00 & 0.00 & 1 & Temperature is 5.0°C above setpoint. Fan must be turned on (1) to reduce temperature, and heater set to 0.0 to prevent further heating. \\
    GPT‑4o      & 25.00 & 20.00 & 20.00 & 0.00 & 0 &  The current temperature T is 5.0C above the setpoint Tsp. Since the ambient temperature Tamb is equal to the setpoint, the natural tendency is for the temperature to decrease towards the setpoint without any intervention. To minimize fan usage, we should avoid using the fan unless necessary. Therefore, we will not use the heater or the fan and allow the temperature to naturally decrease towards the setpoint. \\
    \bottomrule
  \end{tabularx}
  \label{tab:t_above_fan_penalty}
\end{table*}

\begin{table*}[!htbp]
\caption{Comparison of reasoning for respective plans at \(t_{min}=30\) following with heater penalty}
    \centering
    \small
    \setlength{\tabcolsep}{6pt}
    \renewcommand{\arraystretch}{1.3}
    \begin{tabularx}{\textwidth}{l c c c c c X}
    \toprule
    \multirow{2}{*}{Model} & \multicolumn{3}{c}{State} & \multicolumn{2}{c}{Plan} & \multirow{2}{*}{Reasoning} \\
    \cmidrule(lr){2-4}\cmidrule(lr){5-6}
    & $T$ & $T_{sp}$ & $T_{amb}$ & heater & fan & \\
    \midrule
    Gemma-3 1B      & 24.00 & 26.83 & 20.00 & 0.00 & 1 &
    The temperature is below the setpoint, so we need to increase the heater. The fan is already at 20C, so we need to increase the heater's value to bring the temperature up. \\
    Phi-4 Mini 3.8B & 25.51 & 26.83 & 20.00 & 0.01 & 0 &
    The temperature is below the setpoint by 1.9 degrees Celsius and needs to be increased using minimal heater usage while considering fan influence.\\
    Mistral-7B      & 23.86 & 26.83 & 20.00 & 0.00 & 1 &
    To minimize heater usage and approach the setpoint, we will first turn on the fan to reduce the temperature. Since the error (e) is large, a more aggressive action is needed. \\
    Qwen-3 14B      & 26.61 & 26.83 & 20.0 & 0.12 & 0 & Current temperature is 26.0°C (0.4°C below setpoint). The temperature is 0.2°C below the setpoint (e=-0.2). To correct this, the heater should be slightly increased from the previous 0.10 to 0.12, which is a minimal adjustment aligned with the priority of minimal heater usage. The fan remains off to avoid further cooling.\\
    GPT-4o          & 26.77 & 26.83 & 20.00 & 0.00 & 0 & The current temperature T is equal to the setpoint Tsp, resulting in an error e of 0.0C. Since the error is zero, there is no need to adjust the temperature. The previous heater setting was 0.50, which can be reduced to save energy as the temperature is already at the desired setpoint. The fan is off, which is appropriate since we do not need to decrease the temperature further. Therefore, we can turn off the heater to minimize energy usage while maintaining the current temperature.\\
    \bottomrule
\end{tabularx}
\label{tab:reasoning_heater_penalty}
\end{table*}

\begin{table*}[!htbp]
    \caption{Comparison of reasoning for plans when temperature is at setpoint with heater penalty}
    \centering
    \small
    \setlength{\tabcolsep}{6pt}
    \renewcommand{\arraystretch}{1.3}
    \begin{tabularx}{\textwidth}{l c c c c c X}
    \toprule
    \multirow{2}{*}{Model} & \multicolumn{3}{c}{State} & \multicolumn{2}{c}{Plan} & \multirow{2}{*}{Reasoning} \\
    \cmidrule(lr){2-4}\cmidrule(lr){5-6}
                         & $T$ & $T_{sp}$ & $T_{amb}$ & heater & fan &  \\
    \midrule
    Gemma-3 1B   & 25.00 & 25.00 & 20.00 & 0.00 & 0 & The temperature is currently above the setpoint.  We need to decrease the temperature to match the setpoint. \\
    Phi‑4 Mini 3.8B  & 25.00 & 25.00 & 20.00 & 0.00 & 0 & At this minute, the temperature is equal to setpoint with no error (e=0). The fan should be turned off as there’s neither a need for cooling nor heating at equilibrium. \\
    Mistral‑7B  & 25.00 & 25.00 & 20.00 & 0.00 & 0 & The temperature is already at the setpoint and there is no need for the fan or heater. Minimizing heater usage is the first priority. \\
    Qwen‑3 14B  & 25.00 & 25.00 & 20.00 & 0.00 & 0 & The temperature is already at the setpoint (T=25.0C, Tsp=25.0C) with zero error. No heater or fan action is needed to maintain equilibrium. Prioritizing minimal heater usage and avoiding unnecessary fan operation. \\
    GPT‑4o      & 25.00 & 25.00 & 20.00 & 0.00 & 0 & The current temperature T is equal to the setpoint Tsp, resulting in an error e of 0.0C. Since the error is zero, there is no need to adjust the temperature. To minimize energy usage, we should keep the heater off and the fan off as well. \\
    \bottomrule
    \end{tabularx}
    \label{tab:plan_standstill_heater_penalty}
\end{table*}

\begin{table*}[!htbp]
    \caption{Comparison of reasoning for plans when the temperature is below the setpoint with heater penalty}
    \centering
    \small
    \setlength{\tabcolsep}{6pt}
    \renewcommand{\arraystretch}{1.3}
    \begin{tabularx}{\textwidth}{l c c c c c X}
    \toprule
    \multirow{2}{*}{Model} & \multicolumn{3}{c}{State} & \multicolumn{2}{c}{Plan} & \multirow{2}{*}{Reasoning} \\
    \cmidrule(lr){2-4}\cmidrule(lr){5-6}
                         & $T$ & $T_{sp}$ & $T_{amb}$ & heater & fan &  \\
    \midrule
    Gemma-3 1B   & 25.00 & 30.00 & 20.00 & 1.00 & 1 &  The temperature is currently below the setpoint, so we need to increase the temperature.  The heater needs to be turned on to increase the temperature. \\
    Phi‑4 Mini 3.8B  & 25.00 & 30.00 & 20.00 & 0.10 & 0 & Initial temperature is below setpoint; increase heater to warm up while minimizing usage. \\
    Mistral‑7B  & 25.00 & 30.00 & 20.00 & 0.10 & 1 & Initiate the fan to cool down the enclosure and reduce heater usage. \\
    Qwen‑3 14B  & 25.00 & 30.00 & 20.00 & 0.20 & 0 & The temperature is below the setpoint (e=-5.0C), requiring heating. To minimize heater usage while ensuring temperature rises, a low heater value is initiated. The fan is kept off to avoid further cooling. \\
    GPT‑4o      & 25.00 & 30.00 & 20.00 & 0.50 & 0 & The current temperature T is 5.0C below the setpoint Tsp. Since the error |e| is relatively large, we need to increase the temperature efficiently. The heater should be used to raise the temperature, but we should aim to use it minimally. The fan should remain off as it would further decrease the temperature. \\
    \bottomrule
\end{tabularx}
\label{tab:t_below_heater_penalty}
\end{table*}

\begin{table*}[!htbp]
    \caption{Comparison of reasoning for plans when the temperature is above the setpoint with heater penalty}
    \centering
    \small
    \setlength{\tabcolsep}{6pt}
    \renewcommand{\arraystretch}{1.3}
    \begin{tabularx}{\textwidth}{l c c c c c X}
    \toprule
    \multirow{2}{*}{Model} & \multicolumn{3}{c}{State} & \multicolumn{2}{c}{Plan} & \multirow{2}{*}{Reasoning} \\
    \cmidrule(lr){2-4}\cmidrule(lr){5-6}
                         & $T$ & $T_{sp}$ & $T_{amb}$ & heater & fan &  \\
    \midrule
    Gemma-3 1B   & 25.00 & 20.00 & 20.00 & 0.00 & 0 & The temperature is currently above the setpoint.  We need to decrease the temperature to match the setpoint. \\
    Phi‑4 Mini 3.8B  & 25.00 & 20.00 & 20.00 & 0.01 & 1 & Initial temperature is above setpoint; increase fan to cool down gradually while using heater minimally. \\
    Mistral‑7B  & 25.00 & 20.00 & 20.00 & 0.00 & 1 & Initiate the fan to cool down the enclosure as it is currently above the setpoint temperature. The heater will not be used at this moment. \\
    Qwen‑3 14B  & 25.00 & 20.00 & 20.00 & 0.00 & 1 & Current temperature (25.0C) is 5.0C above setpoint (20.0C). Need to cool down. Fan should be turned on (1) to reduce temperature. Heater must remain off (0.0) to minimize usage. \\
    GPT‑4o      & 25.00 & 20.00 & 20.00 & 0.00 & 1 & The current temperature T is 5.0C above the setpoint Tsp. To reduce the temperature, we should use the fan. Since the error is significant, we can turn the fan on fully. The heater should remain off to minimize energy usage. \\
    \bottomrule
\end{tabularx}
\label{tab:t_above_heater_penalty}
\end{table*}

% Reaosning table with and without prediction for Qwen
\begin{table*}[!htbp]
  \caption{Comparison of reasoning for Qwen-3 14B and GPT-4o with and without prediction tool}
  \centering
  \small
  \setlength{\tabcolsep}{6pt}
  \renewcommand{\arraystretch}{1.2}
  \begin{tabularx}{\textwidth}{l c c c c c X}
    \toprule
    \multirow{2}{*}{Model} & \multicolumn{3}{c}{State} & \multicolumn{2}{c}{Plan} & \multirow{2}{*}{Reasoning} \\
    \cmidrule(lr){2-4}\cmidrule(lr){5-6}
                           & $T$ & $T_{sp}$ & $T_{amb}$ & heater & fan & \\
    \midrule
    % ---- Qwen-3 14B ----
    \multirow[t]{10}{*}{Qwen-3 14B} &
    \multirow[t]{10}{*}{25.93} &
    \multirow[t]{10}{*}{26.74} &
    \multirow[t]{10}{*}{20.00} &
    0.82 & 0 &
    \multirow[t]{10}{=}{The current temperature matches the setpoint (T=25.9°C, Tsp=25.9°C), but the setpoint will increase gradually over the next 10 minutes. 
    To follow the rising setpoint, the heater must be increased incrementally while keeping the fan off (to avoid cooling). 
    Starting with a small increase from the previous heater value (0.80) and gradually raising it by 0.02 per minute ensures a smooth temperature rise without overshooting. The fan remains off throughout to maximize heating efficiency.} \\
    & & & & 0.84 & 0 & \\
    & & & & 0.86 & 0 & \\
    & & & & 0.88 & 0 & \\
    & & & & 0.90 & 0 & \\
    & & & & 0.92 & 0 & \\
    & & & & 0.94 & 0 & \\
    & & & & 0.96 & 0 & \\
    & & & & 0.98 & 0 & \\
    & & & & 1.00 & 0 & \\
    \midrule
    % ---- Qwen-3 14B (tool on) ----
    \multirow[t]{10}{*}{Qwen-3 14B tool} &
    \multirow[t]{10}{*}{25.79} &
    \multirow[t]{10}{*}{26.74} &
    \multirow[t]{10}{*}{20.00} &
    0.55 & 0 &
    \multirow[t]{10}{=}{Based on the first simulation's final error of 1.16C, the second plan with fan activation is likely to track the setpoint better. The fan was turned on earlier to counteract potential overshoot, and heater values were adjusted to avoid excessive temperature increases.} \\
    & & & & 0.60 & 0 & \\
    & & & & 0.65 & 0 & \\
    & & & & 0.70 & 1 & \\
    & & & & 0.75 & 1 & \\
    & & & & 0.80 & 1 & \\
    & & & & 0.85 & 0 & \\
    & & & & 0.90 & 0 & \\
    & & & & 0.95 & 0 & \\
    & & & & 1.00 & 0 & \\
    \bottomrule
    % ---- GPT-4o ----
    \multirow[t]{10}{*}{GPT-4o} &
    \multirow[t]{10}{*}{28.21} &
    \multirow[t]{10}{*}{26.74} &
    \multirow[t]{10}{*}{20.00} &
    0.50 & 1 &
    \multirow[t]{10}{=}{The current temperature T is 28.2°C, which is 1.4°C above the current setpoint Tsp of 26.8°C. The upcoming setpoints are gradually increasing to 27.0°C. To reduce the temperature towards the setpoint, we should use the fan to cool down the enclosure. Since the error |e| is moderate, we can use the fan consistently and reduce the heater to zero to allow the temperature to decrease. As the setpoint increases slightly, 
we will adjust the heater to maintain the temperature close to the setpoint.} \\
    & & & & 0.50 & 1 & \\
    & & & & 0.50 & 1 & \\
    & & & & 0.50 & 1 & \\
    & & & & 0.50 & 1 & \\
    & & & & 0.50 & 1 & \\
    & & & & 0.50 & 1 & \\
    & & & & 0.50 & 1 & \\
    & & & & 0.50 & 1 & \\
    & & & & 0.50 & 1 & \\
    \bottomrule
    % ---- GPT-4o (tool on) ----
    \multirow[t]{10}{*}{GPT-4o} &
    \multirow[t]{10}{*}{26.85} &
    \multirow[t]{10}{*}{26.74} &
    \multirow[t]{10}{*}{20.00} &
    0.30 & 1 &
    \multirow[t]{10}{=}{The final plan balances the heater and fan usage to closely follow the setpoint changes, resulting in a final error of +0.34°C. This approach prevents the temperature from dropping too much initially and stabilizes it around the setpoint in the later steps.} \\
    & & & & 0.40 & 0 & \\
    & & & & 0.40 & 1 & \\
    & & & & 0.40 & 0 & \\
    & & & & 0.40 & 0 & \\
    & & & & 0.40 & 0 & \\
    & & & & 0.40 & 0 & \\
    & & & & 0.40 & 0 & \\
    & & & & 0.40 & 0 & \\
    & & & & 0.40 & 0 & \\
    \bottomrule
  \end{tabularx}
\end{table*}

% The 2 simulations Qwen-3 14b ran before deciding on a plan
\begin{toolbox}
\textbf{Sim1:} Qwen-3 14B calls \texttt{simulate\_plan}.
\begin{lstlisting}[style=sqlstyle] 
t+01: T=25.79->25.94C, e_end=+0.05C, uQ=0.60, uF=0
t+02: T=25.94->26.11C, e_end=+0.13C, uQ=0.65, uF=0
t+03: T=26.11->26.29C, e_end=+0.21C, uQ=0.70, uF=0
t+04: T=26.29->26.48C, e_end=+0.31C, uQ=0.75, uF=0
t+05: T=26.48->26.69C, e_end=+0.42C, uQ=0.80, uF=0
t+06: T=26.69->26.91C, e_end=+0.55C, uQ=0.85, uF=0
t+07: T=26.91->27.14C, e_end=+0.68C, uQ=0.90, uF=0
t+08: T=27.14->27.39C, e_end=+0.84C, uQ=0.95, uF=0
t+09: T=27.39->27.64C, e_end=+1.00C, uQ=1.00, uF=0
t+10: T=27.64->27.90C, e_end=+1.16C, uQ=1.00, uF=0
Predicted final error: +1.16C
\end{lstlisting}
\end{toolbox}

\begin{toolbox}
\textbf{Sim2:} Qwen-3 14B calls \texttt{simulate\_plan}.
\begin{lstlisting}[style=sqlstyle] 
t+01: T=25.79->25.94C, e_end=+0.05C, uQ=0.60, uF=0
t+02: T=25.94->26.11C, e_end=+0.13C, uQ=0.65, uF=0
t+03: T=26.11->26.29C, e_end=+0.21C, uQ=0.70, uF=0
t+04: T=26.29->26.48C, e_end=+0.31C, uQ=0.75, uF=0
t+05: T=26.48->26.44C, e_end=+0.17C, uQ=0.80, uF=1
t+06: T=26.44->26.41C, e_end=+0.05C, uQ=0.85, uF=1
t+07: T=26.41->26.39C, e_end=-0.06C, uQ=0.90, uF=1
t+08: T=26.39->26.64C, e_end=+0.09C, uQ=0.95, uF=0
t+09: T=26.64->26.89C, e_end=+0.25C, uQ=1.00, uF=0
t+10: T=26.89->27.15C, e_end=+0.41C, uQ=1.00, uF=0
Predicted final error: +0.41C
\end{lstlisting}
\end{toolbox}

% The 3 simulations it GPT-4o ran before deciding on a plan
\begin{toolbox}
\textbf{Sim1:} GPT-4o calls \texttt{simulate\_plan}.
\begin{lstlisting}[style=sqlstyle] 
t+01: T=26.85->26.93C, e_end=+0.10C, uQ=0.30, uF=0
t+02: T=26.93->27.03C, e_end=+0.10C, uQ=0.40, uF=0
t+03: T=27.03->27.16C, e_end=+0.16C, uQ=0.50, uF=0
t+04: T=27.16->27.29C, e_end=+0.29C, uQ=0.50, uF=0
t+05: T=27.29->27.42C, e_end=+0.42C, uQ=0.50, uF=0
t+06: T=27.42->27.55C, e_end=+0.55C, uQ=0.50, uF=0
t+07: T=27.55->27.68C, e_end=+0.68C, uQ=0.50, uF=0
t+08: T=27.68->27.81C, e_end=+0.81C, uQ=0.50, uF=0
t+09: T=27.81->27.94C, e_end=+0.94C, uQ=0.50, uF=0
t+10: T=27.94->28.06C, e_end=+1.06C, uQ=0.50, uF=0
Predicted final error: +1.06C
\end{lstlisting}
\end{toolbox}

\begin{toolbox}
\textbf{Sim2:} GPT-4o calls \texttt{simulate\_plan}.
\begin{lstlisting}[style=sqlstyle] 
t+01: T=26.85->26.64C, e_end=-0.19C, uQ=0.20, uF=1
t+02: T=26.64->26.46C, e_end=-0.46C, uQ=0.30, uF=1
t+03: T=26.46->26.32C, e_end=-0.68C, uQ=0.40, uF=1
t+04: T=26.32->26.18C, e_end=-0.82C, uQ=0.40, uF=1
t+05: T=26.18->26.05C, e_end=-0.95C, uQ=0.40, uF=1
t+06: T=26.05->25.92C, e_end=-1.08C, uQ=0.40, uF=1
t+07: T=25.92->25.79C, e_end=-1.21C, uQ=0.40, uF=1
t+08: T=25.79->25.67C, e_end=-1.33C, uQ=0.40, uF=1
t+09: T=25.67->25.56C, e_end=-1.44C, uQ=0.40, uF=1
t+10: T=25.56->25.45C, e_end=-1.55C, uQ=0.40, uF=1
Predicted final error: -1.55C
\end{lstlisting}
\end{toolbox}

\begin{toolbox}
\textbf{Sim3:} GPT-4o calls \texttt{simulate\_plan}.
\begin{lstlisting}[style=sqlstyle] 
t+01: T=26.85->26.67C, e_end=-0.17C, uQ=0.30, uF=1
t+02: T=26.67->26.77C, e_end=-0.16C, uQ=0.40, uF=0
t+03: T=26.77->26.61C, e_end=-0.39C, uQ=0.40, uF=1
t+04: T=26.61->26.72C, e_end=-0.28C, uQ=0.40, uF=0
t+05: T=26.72->26.82C, e_end=-0.18C, uQ=0.40, uF=0
t+06: T=26.82->26.92C, e_end=-0.08C, uQ=0.40, uF=0
t+07: T=26.92->27.03C, e_end=+0.03C, uQ=0.40, uF=0
t+08: T=27.03->27.13C, e_end=+0.13C, uQ=0.40, uF=0
t+09: T=27.13->27.23C, e_end=+0.23C, uQ=0.40, uF=0
t+10: T=27.23->27.34C, e_end=+0.34C, uQ=0.40, uF=0
Predicted final error: +0.34C
\end{lstlisting}
\end{toolbox}

\end{document}